%% file: supratex.tex
\documentclass[10pt]{article}
\usepackage{supratex}

\input{math_commands.tex}

\usepackage{longtable,float,fvextra,placeins}
\newcolumntype{L}[1]{>{\raggedright\arraybackslash}p{#1}}
\definecolor{texsplit}{HTML}{173B63}
\newcommand{\splitname}[1]{\textcolor{texsplit}{\textit{#1}}}
\input{commands.tex}
\input{authors.tex}

\title{SUPRA-TEX:}
\suprasubtitle{3D Generation with Super-Aligned Texture}
\suprashorttitle{\method}
\supraabstract{\input{abstract.tex}}
\suprateaser{%
  \includegraphics[width=\linewidth,page=1]{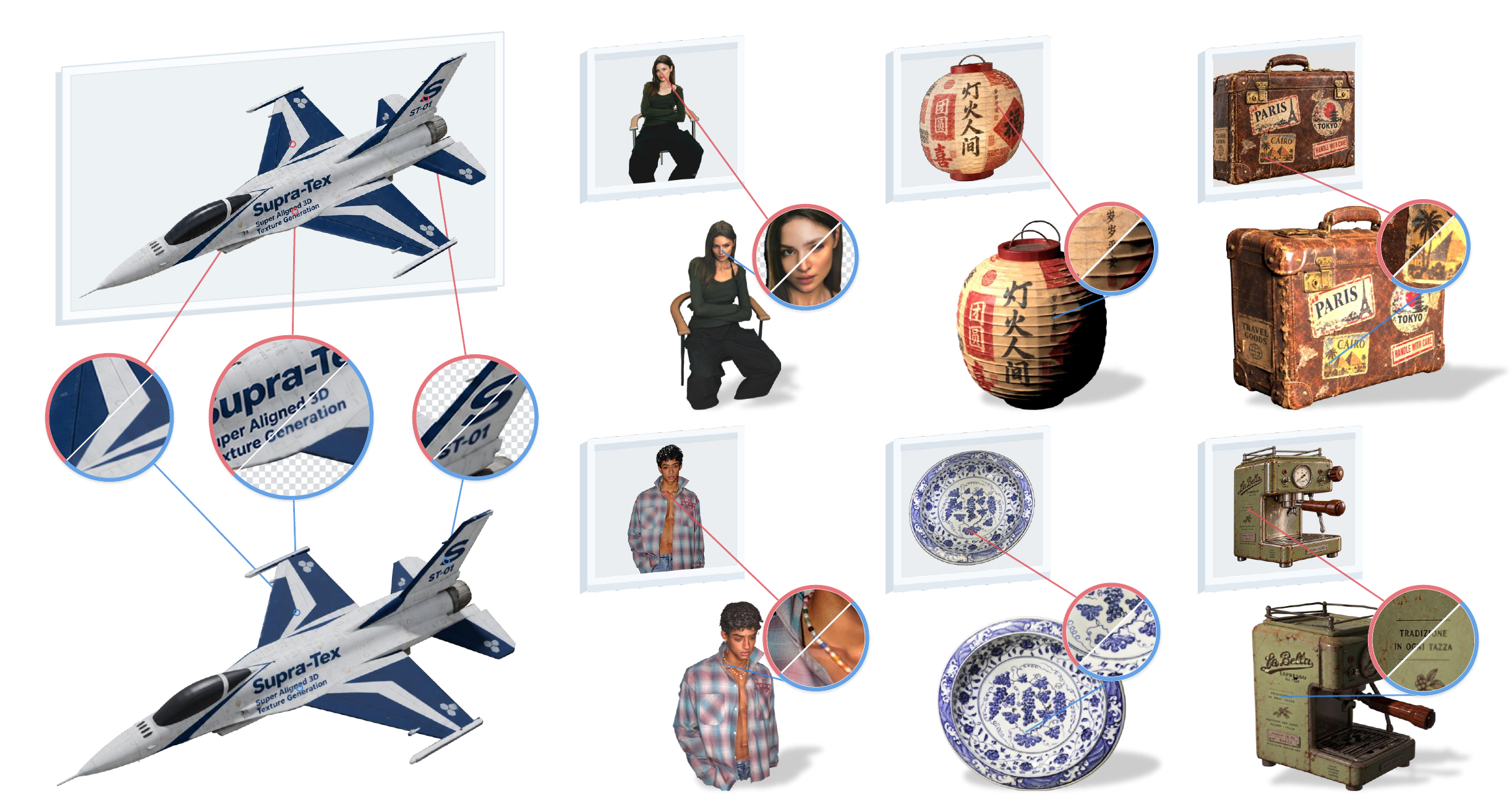}%
}{Super-aligned textured mesh generated by \method{}. Our method preserves fine-grained appearance details from the reference image. \textbf{\emph{(Best viewed with zoom.)}}}{fig:teaser}

\begin{document}
\maketitle
\clearpage
\section{Introduction}
\label{sec:introduction}

For applications such as product digitization and embodied AI, generating relightable 3D assets from a single reference image requires faithful preservation of the reference appearance. We refer to this requirement as \emph{super-alignment}: accurately reproducing fine-grained text, logos, and patterns at their corresponding visible surface locations, while completing unobserved regions with materials consistent with the visible appearance.

Two prominent approaches to appearance generation are multi-view synthesis and native 3D generation. Multi-view pipelines leverage geometry-conditioned 2D generative models to synthesize appearance across surrounding viewpoints, then project and fuse the resulting images onto the mesh~\citep{hunyuan3d2025hunyuan3d}. However, cross-view discrepancies can introduce seams and ghosting, while blending conflicting predictions blurs sharp reference details~\citep{lai2026natex,zeng2025textrix}.

Native 3D generation circumvents view-fusion artifacts by synthesizing appearance directly on 3D representations~\citep{lai2026natex,chen2026lafite,zeng2025textrix,xiang2026native}, yet preserving fine reference details remains challenging. For example, TRELLIS.2 employs a 3D material VAE to compress surface PBR properties into material latents, which are generated under geometry and image conditioning before being decoded into surface materials. Its image cross-attention provides no explicit correspondence between image details and 3D surface locations, leaving the generator to learn where those details belong. To improve spatial alignment, recent work lifts high-resolution image features into 3D through camera projection~\citep{li2026pixal3d}. Yet even with explicit spatial correspondence, these conditions remain visual features rather than material observations. Consequently, the 3D generator must simultaneously shoulder the responsibilities of inferring PBR properties from visual features, reconstructing fine reference details at their corresponding surface locations, and synthesizing materials for unobserved regions.

We argue that these responsibilities should not fall on a single 3D generator. Given the inherent scarcity of 3D training data, requiring the 3D model to learn these heterogeneous tasks jointly increases its learning burden. Instead, we offload visible-material estimation to 2D generative priors and establish explicit surface alignment through camera projection, separating both from the 3D generative task. Our key insight is to translate visible reference evidence into \emph{native material observations} before 3D generation, then organize full-surface completion around this grounded evidence. Rather than forcing the 3D model to infer physical materials from abstract visual features, this formulation enables the generator to anchor directly on established material evidence, preserving reference details while focusing more of its capacity on completing unobserved surfaces.

Based on this principle, we introduce \suprahighlight{SUPRA-Tex}, a framework for generating relightable 3D assets with super-aligned textures. First, we leverage strong 2D priors by training an MLLM-based image editing model to jointly predict lighting-separated PBR maps in the reference view in a single forward pass, while preserving fine visual details. Next, we project these maps onto valid visible surface locations and encode the resulting partial materials with the pretrained material VAE encoder into \emph{native observation latents}. Sharing the target latent space and directly decodable by the pretrained material decoder, these latents provide spatially aligned material content rather than abstract visual features that the 3D backbone must translate into PBR properties. We also encode the spatial extent of these observations to guide completion.

To guide full-surface material generation, we design an \emph{interleaved flow transformer} that reads the native observation latents as fixed, read-only reference memory at every block. Its blocks alternate between updating unobserved latent cells and the full surface, allocating more updates to cells without observations while retaining full-surface interactions for coherent completion. Since observations are indexed by 3D surface coordinates, the same interface incorporates additional views, including 2D-generated back views, and interactive 2D edits without retraining the 3D backbone. Through this division of labor, \method{} delivers \emph{What You See Is What You Get} fidelity in the reference view while completing the asset with globally coherent, relightable materials, as illustrated in Fig.~\ref{fig:teaser}.

Using only public datasets for training, \method{} achieves SOTA results in both geometry-conditioned texturing and end-to-end asset generation. With synthetic back-view conditioning, \method{} achieves reference fidelity and visual quality on real photographs comparable to or better than those of the leading commercial systems in our comparisons. 
Code and checkpoints are released on our project page to facilitate future research: \href{\projectpageurl}{\texttt{supra-3d.github.io/supratex}}.

Our main contributions are as follows:

\begin{itemize}

  \item Our key insight is to establish surface-aligned material observations before 3D generation, allowing the generator to focus on preserving reference details and completing unobserved surfaces. Built on this insight, \method{} is trained solely on public datasets and, with synthetic back-view conditioning, achieves reference fidelity and visual quality comparable to or better than those of leading commercial systems in our comparisons.

  \item We present \emph{native observation encoding}, which projects reference-view PBR estimates onto valid surface locations and encodes partial materials in the target latent space, simplifying learning to preserve reference appearance.

  \item We propose an \emph{interleaved flow transformer} that uses fixed, read-only observation anchors and alternates unobserved-cell and full-surface updates for reference-faithful, coherent material completion.

\end{itemize}

\clearpage
\section{Related Work}
\label{sec:related}

\noindent\textbf{Multi-view texture generation.}
Most texturing pipelines synthesize appearance in 2D views and fuse the predictions onto the mesh. Early approaches paint the mesh view-by-view with depth-guided diffusion and inpaint unobserved UV regions~\citep{richardson2023texture,chen2023text2tex,zeng2024paint3d}. TexFusion aggregates view-wise denoising predictions in a shared latent texture map~\citep{cao2023texfusion}. Subsequent works improve cross-view coherence by synchronizing multi-view denoising~\citep{liu2024syncmvd,cheng2025mvpaint,bensadoun2024texturegen} or generating views jointly via geometry-conditioned diffusion~\citep{huang2024mvadapter,shao2025mvpainter,liang2025unitex,han2026ink3d,zhang2026pixtex}, a recipe also adopted by large image-to-3D systems~\citep{zhang2024clay,li2025triposg,zhao2025hunyuan3d2}.
Reference-guided methods also explicitly target appearance preservation: EASI-Tex combines mesh-edge guidance with image features for single-image texturing~\citep{perla2024easitex}, while AlignTex aligns multi-view artwork with the mesh before projection and texture refinement~\citep{zhang2025aligntex}.
Optimization-based approaches such as DreamMat distill geometry- and light-aware diffusion priors into PBR materials~\citep{zhang2024dreammat}. Paint-it instead optimizes convolutionally parameterized PBR texture maps with score distillation~\citep{youwang2024paintit}.
To produce relightable assets, multi-view generation is extended to PBR channels using lighting-aware priors, material latent representations, or alternating view-space baking~\citep{deng2024flashtex,huang2025materialanything,he2025materialmvp,yeo2026matlat,feng2025romantex,hunyuan3d2025hunyuan3d,lai2025hunyuan3d25,bao2025lumitex,wu2026matmart}, often aided by single-image intrinsic delighting~\citep{kocsis2024intrinsic,zeng2024rgbx,ke2025marigold,chen2024intrinsicanything,he2025neurallightrig,careaga2024colorful}. Nevertheless, residual inconsistencies between synthesized views can blur or ghost sharp reference details during projection and blending. In contrast, \method{} completes materials in native 3D space, using projected observations as conditioning anchors rather than directly fusing synthesized views into the final texture.

\noindent\textbf{Native texture generation.}
Driven by the rise of large-scale 3D pretraining, native texture generation has emerged as an alternative paradigm. Existing approaches explore diffusion over point, octree, and UV representations~\citep{yu2023pointuv,liu2024texoct,yu2024texgen} and PBR prediction with octree-aligned Gaussians~\citep{xiong2025texgaussian}. Latent texture generators, such as NaTex, LaFiTe, and TEXTRIX, learn structured latent spaces for surface colors or attributes~\citep{lai2026natex,chen2026lafite,zeng2025textrix}, while native 3D foundation models jointly represent geometry and appearance~\citep{chen2025threedtopiaxl,xiang2025structured,xiang2026native}. Beyond global conditioning, partial observations and spatially aligned features have also been explored. At scene scale, WorldGrow~\citep{li2026worldgrow} extends TRELLIS with context-conditioned 3D block inpainting for progressive scene generation. TEXGen back-projects partial UV maps~\citep{yu2024texgen}, NaTex encodes initial textures for refinement~\citep{lai2026natex}, and TEXTRIX encodes projected appearance with its pretrained attribute VAE in the generation target's latent space~\citep{zeng2025textrix}. For spatial feature alignment, camera projection techniques originally developed for implicit reconstruction~\citep{saito2019pifu,yu2021pixelnerf} have been adapted to 3D generative models, such as CUPID's pose-aligned features and Pixal3D's camera-frame back-projected multi-scale features~\citep{huang2025cupid,li2026pixal3d}. Even with spatially aligned image features, the 3D generator must still infer material properties, preserve reference details, and complete unobserved surfaces.

\section{Method}
\label{sec:method}

\begin{figure*}[!t]
  \centering
  \includegraphics[width=\linewidth,trim=0bp 5.01bp 0bp 6.72bp,clip]{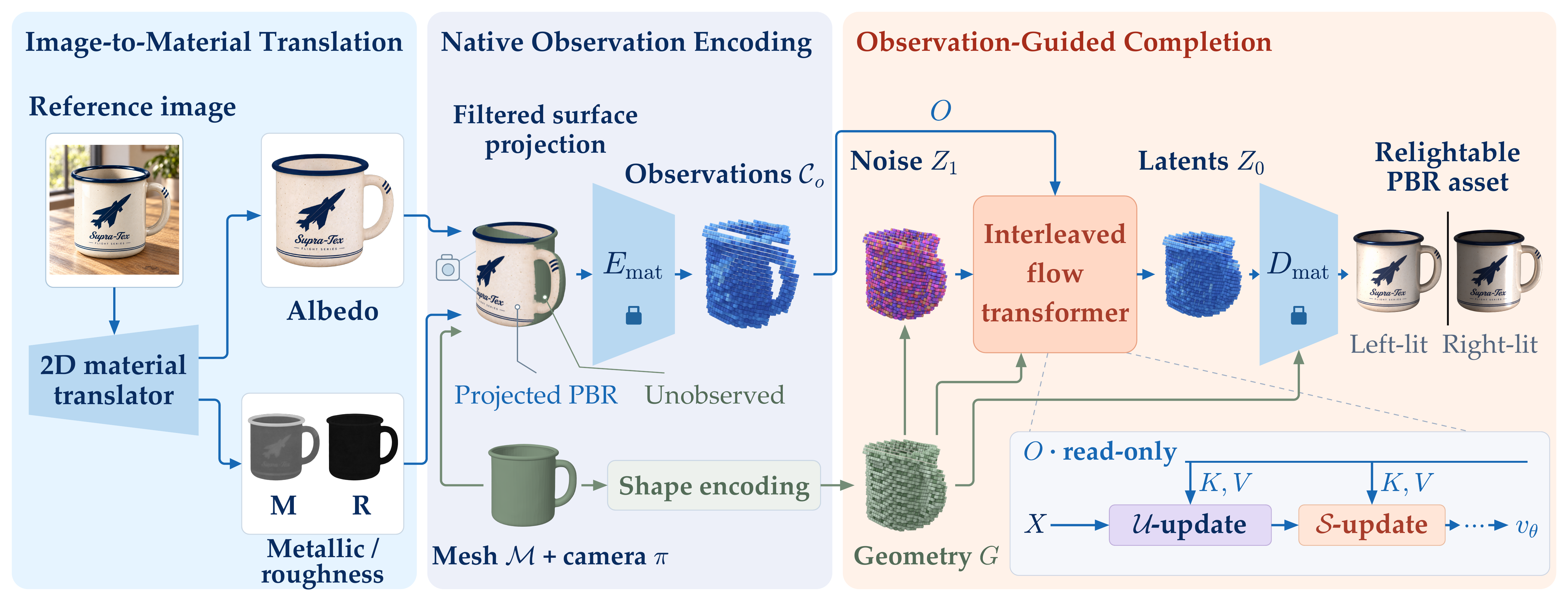}
  \caption{\textbf{Overview of \method{}.}
  \textbf{Left: image-to-material translation.} A 2D material translator estimates lighting-separated albedo, metallic, and roughness maps from a single reference image.
  \textbf{Middle: native observation encoding.} Using the mesh $\mathcal M$ and camera projection $\pi$, we project valid predictions onto visible surface voxels and omit unobserved voxels from the partial PBR input. The frozen material encoder $E_{\mathrm{mat}}$ maps these partial PBR attributes into the generation target's latent space. These latents, together with coverage and boundary metadata, form the observation representation $\mathcal C_o$.
  \textbf{Right: observation-guided completion.} Conditioned on geometry latents $G$, the interleaved flow transformer generates complete material latents $Z_0$ from noise $Z_1$. In the inset, the fixed, read-only stream $O$ supplies keys and values, while the generation stream $X$ alternates updates over cells without observations ($\mathcal U$) and the full surface ($\mathcal S$). The frozen decoder $D_{\mathrm{mat}}$ converts $Z_0$ into surface PBR materials, lit from the left and right.}
  \label{fig:overview}
\end{figure*}

Given a reference image $I$, a mesh $\mathcal M$, and its camera projection $\pi$, \method{} generates full-surface, relightable PBR materials. Figure~\ref{fig:overview} summarizes the three stages: reference-view material estimation, surface-aligned native observation encoding, and observation-guided 3D completion. We first introduce the material representation and upstream geometry model, then describe the three stages, training objective, and inference procedure. Appendices~\ref{app:method-details} and~\ref{app:training-details} provide implementation and training details; Appendix~\ref{app:texture-editing} describes additional-view conditioning and texture editing.

\subsection{Preliminaries}
\label{sec:preliminaries}

\noindent\textbf{Surface PBR and material latents.}
We represent surface appearance through physically based rendering (PBR) properties over the mesh $\mathcal M$. At any surface location $\mathbf p \in \mathcal M$, these properties comprise albedo $\mathbf a\in\mathbb R^3$, metallic $m\in[0,1]$, and roughness $r\in[0,1]$. We build on the O-Voxel representation of TRELLIS.2~\citep{xiang2026native}, which additionally includes opacity $\alpha\in[0,1]$, giving the material tuple $T(\mathbf p)=(\mathbf a(\mathbf p),m(\mathbf p),r(\mathbf p),\alpha(\mathbf p))$. Separate sparse VAEs compress 3D geometry and surface materials into geometry latents $G=\{g_s\}_{s\in\mathcal S}$ and material latents $Z=\{z_s\}_{s\in\mathcal S}$ over active coordinates $\mathcal S$. For sparse surface attributes $\mathcal P$, the pretrained material VAE encoder $E_{\mathrm{mat}}$ yields $Z$; conditioned on $G$, the material generation stage models PBR materials directly in the native 3D space by generating $Z$, which the decoder $D_{\mathrm{mat}}$ restores onto the mesh surface.

\noindent\textbf{Geometry foundation and camera projection.}
Our framework takes a mesh $\mathcal M$ and camera projection $\pi$, either provided as ground truth or predicted from the reference image. We use the open-source TRELLIS.2-based Pixal3D variant~\citep{li2026pixal3d,xiang2026native}, denoted Pixal3D$^\star$, which supports geometry and material generation. Its geometry module predicts the reference-aligned mesh for end-to-end generation.

\subsection{Image-to-Material Translation}
\label{sec:translation}

To reduce the learning burden on the 3D generator, we separate reference-view material estimation from 3D surface completion. Specifically, we train an MLLM-based image editing model $F_{\mathrm{2D}}$ to infer reference-view albedo and metallic-roughness (MR) maps directly from the input photograph:
\begin{equation}
\widehat T_{\mathrm{2D}}=F_{\mathrm{2D}}(I)
=\left(\widehat{\mathbf a},\widehat m,\widehat r\right),
\label{eq:material-translation}
\end{equation}
where $\widehat{\mathbf a}$ denotes the albedo map, and $\widehat m$ and $\widehat r$ denote the metallic and roughness maps. To reuse the 3-channel pretrained image VAE, we pack $\widehat m$ and $\widehat r$ with a constant channel into a 3-channel map, which also fulfills the input format expected by the pretrained material encoder.

\noindent\textbf{Joint alignment and one-step inference.}
To coordinate channel predictions while retaining pretrained image priors, $F_{\mathrm{2D}}$ adapts a multimodal DiT backbone~\citep{firered2026imageedit} via \emph{factorized modality-axis joint attention}, jointly predicting albedo, metallic--roughness, and auxiliary normal maps. This attention exchanges features across output modalities at corresponding spatial locations. We train the 2D material estimator on RGB images paired with ground-truth material and normal maps rendered from Objaverse, using a latent flow-matching objective at the fixed maximum-noise endpoint $\sigma=1.0$. At inference, a single DiT forward pass at this endpoint (1.2\,s on one H800) estimates clean material latents, which the VAE decodes into $1024\times1024$ material maps without iterative denoising.

This design has two advantages. First, it uses 2D image priors to disentangle illumination from intrinsic materials at full reference resolution. Second, it shields the downstream 3D generator from abstract visual features, letting the 3D model use physical PBR observations to focus on reference preservation and unobserved-surface completion.

\subsection{Native Observation Encoding}
\label{sec:alignment}

To supply physical observations to the 3D generator in a representation it natively understands, we bridge the estimated reference-view PBR maps $\widehat T_{\mathrm{2D}}$ directly to the surface of $\mathcal M$ using the pretrained material encoder $E_{\mathrm{mat}}$.

\noindent\textbf{Filtered surface projection.}
Through the camera projection $\pi$, we associate each foreground pixel $\mathbf u$ of $\widehat T_{\mathrm{2D}}$ with its first visible surface hit $\mathbf p \in \mathcal M$. At inference, foreground-mask erosion and local depth-discontinuity filtering further remove unreliable boundary correspondences. The retained correspondences form the observation set $\mathcal H_{\mathrm{valid}}$, shared by all PBR channels. We voxelize the retained materials into a partial surface representation:
\begin{equation}
\mathcal P_o=
\operatorname{Voxelize}\!\left(
\left\{\left(\mathbf p,\widehat T_{\mathrm{2D}}(\mathbf u)\right)
\mid (\mathbf u,\mathbf p)\in\mathcal H_{\mathrm{valid}}\right\}
\right),
\label{eq:local-pbr}
\end{equation}
where $\mathbf u$ is a 2D pixel coordinate and $\mathbf p$ its surface location on $\mathcal M$; $\mathcal P_o$ omits regions without valid observations.

\noindent\textbf{Sparse coordinate encoding.}
The partial surface attributes $\mathcal P_o$ match the input domain of the pretrained material VAE, allowing direct ingestion by $E_{\mathrm{mat}}$ without intermediate adapters:
\begin{equation}
E_{\mathrm{mat}}(\mathcal P_o)
=\left\{(s,z_s^o)\mid s\in\mathcal S_o\right\},
\qquad \mathcal S_o\subseteq\mathcal S,
\label{eq:observation-encoding}
\end{equation}
using the posterior mean. We call $\{z_s^o\}$ \emph{native observation latents}; $\mathcal S_o$ indexes latent cells receiving observations.

Zero-padding unobserved regions assigns concrete material values, such as black albedo, to surfaces without observations. We instead use the sparse coordinate architecture of $E_{\mathrm{mat}}$ to omit unobserved voxels. Encoding only the visible support $\mathcal P_o$ avoids assigning explicit material values to unobserved voxels while retaining the visible support as conditioning evidence. This encoding runs once before sampling and remains fixed throughout generation. Section~\ref{sec:observation-analysis} examines how its latents and decoded materials differ from complete-material encoding.

\noindent\textbf{Spatial observation metadata.}
Each latent cell aggregates surface voxels, so an observed cell in $\mathcal S_o$ is rarely fully covered from one view. The generator must quantify each cell's observation support and proximity to unobserved regions. Let $\mathcal V_s$ be the surface voxels in cell $s$ and $\mathcal V_o$ the filtered, deduplicated set of observed voxels. Coverage is
\begin{equation}
\rho_s=\frac{|\mathcal V_s\cap\mathcal V_o|}{|\mathcal V_s|},
\qquad s\in\mathcal S_o,
\label{eq:coverage}
\end{equation}
preventing repeated pixel projections onto one voxel from inflating coverage. Let $\mathcal U=\mathcal S\setminus\mathcal S_o$ be the unobserved cells and $\mathcal G_{\mathcal S}$ the six-neighbor graph over $\mathcal S$; with $\mathcal U_s\subseteq\mathcal U$ the unobserved cells reachable from $s$, the boundary distance is
\begin{equation}
d_s=
\begin{cases}
\displaystyle\min_{q\in\mathcal U_s}
\operatorname{dist}_{\mathcal G_{\mathcal S}}(s,q),
&\mathcal U_s\neq\varnothing,\\
-1,&\mathcal U_s=\varnothing,
\end{cases}
\label{eq:boundary-distance}
\end{equation}
measured in graph steps, with $-1$ marking cells from which no unobserved region is reachable. Both quantities describe the extent of observation, not the accuracy of the 2D estimate. These native observation latents and their spatial extent form the native observation representation
\begin{equation}
\mathcal C_o=
\left\{(s,z_s^o,\rho_s,d_s)\mid s\in\mathcal S_o\right\},
\label{eq:observation-condition}
\end{equation}
which encapsulates native material observations alongside their physical extent without fictitious padding, ready to anchor full-surface generation. Because anchors are indexed in 3D coordinates, the same interface can combine observations from additional or edited views without retraining the generator (Appendix~\ref{app:texture-editing}).

\subsection{Observation-Guided Completion}
\label{sec:completion}

We generate complete material latents via flow matching over the mesh $\mathcal M$, predicting a velocity field at time $t$ from the current noisy material latents $Z_t$, geometry latents $G$, and native observation representation $\mathcal C_o$~\citep{lipman2023flow}. Native observation latents provide spatially aligned material evidence for reconstructing visible appearance. The generator must learn to use this evidence while completing unobserved surfaces coherently. Pixal3D$^\star$ injects projected image features through additive residuals and updates all surface tokens at every layer. We introduce an interleaved flow transformer that retains observations as fixed, read-only memory and alternates unobserved-cell and full-surface updates. This schedule assigns more hidden-state updates to unobserved cells while allowing visible materials to be refined during full-surface generation.

\noindent\textbf{Interleaved flow transformer.}
The generator uses two coordinate supports: the full active surface $\mathcal S$ and the observed subset $\mathcal S_o \subseteq \mathcal S$. Figure~\ref{fig:interleaved-flow} in Appendix~\ref{app:transformer-details} illustrates the update schedule. We maintain two streams: a dynamic generation stream $X$ tracking the hidden state of $Z_t$ over $\mathcal S$, and a static observation stream $O$ serving as immutable reference memory over $\mathcal S_o$. At observed coordinates, $O_s$ stores partial evidence while $X_s$ predicts the complete cell. Both streams use the same 3D coordinate frame and a shared input projection $W_{\mathrm{in}}$ over material and geometry latents:
\begin{equation}
\begin{aligned}
X_s^{(0)}&=W_{\mathrm{in}}([Z_{t,s};g_s]),
&&s\in\mathcal S,\\
O_s&=W_{\mathrm{in}}([\bar z_s^o;g_s])
+W_{\mathrm{meta}}([\rho_s;d_s])+b_o,
&&s\in\mathcal S_o,
\end{aligned}
\label{eq:stream-inputs}
\end{equation}
where $\bar z_s^o$ scales the native observation latents to match the normalization of $Z_t$, $W_{\mathrm{meta}}$ is a learnable affine projection of spatial metadata, and $b_o$ is a learned observation-role bias initialized to zero.

Across the Transformer blocks, $O$ produces no queries and receives no updates with $O^{(\ell+1)} = O$ across blocks, serving as read-only memory supplying keys and values to every layer. To concentrate generative capacity on unobserved surfaces, query coordinates alternate across blocks $\ell = 0, \ldots, L-1$: $\mathcal J_\ell = \mathcal U$ for even $\ell$ over unobserved cells $\mathcal U = \mathcal S \setminus \mathcal S_o$, and $\mathcal J_\ell = \mathcal S$ for odd $\ell$ over the full surface to maintain global coherence. Let $\widetilde X^{(\ell)}=N_\ell(X^{(\ell)},t)$ and $\widetilde O^{(\ell)}=N_\ell(O,0)$ denote the streams after normalization and time modulation, where observations carry a fixed clean-time indicator. The cross-stream attention is formulated as:
\begin{equation}
\begin{aligned}
Q_\ell = q_\ell\!\left(\widetilde X^{(\ell)}_{\mathcal J_\ell}\right), \quad
K_\ell = \left[k_\ell\!\left(\widetilde O^{(\ell)}\right); k_\ell\!\left(\widetilde X^{(\ell)}\right)\right], \quad
V_\ell = \left[v_\ell\!\left(\widetilde O^{(\ell)}\right); v_\ell\!\left(\widetilde X^{(\ell)}\right)\right], \\
A_\ell = \operatorname{softmax}\!\left(Q_\ell K_\ell^\top / \sqrt{d_h}\right) V_\ell,
\end{aligned}
\label{eq:attention}
\end{equation}
where $d_h=128$ is the per-head dimension and $q_\ell,k_\ell$ apply 3D RoPE to latent coordinates, enabling attention based on relative position and content. Residual updates affect only $\mathcal J_\ell$; other generation tokens are unchanged in that block, while $O$ remains fixed throughout. An output projection maps the terminal state $X^{(L)}$ to predicted velocities for all of $\mathcal S$. The full-surface training objective and sampling procedure are given in Sec.~\ref{sec:training-inference}.

\subsection{Training and Inference}
\label{sec:training-inference}

\noindent\textbf{Full-surface training objective.}
The generation stream $X$ synthesizes material latents over the surface $\mathcal S$. Training observations are constructed from ground-truth materials of public 3D assets using visible-surface observation encoding, with complete material latents as targets. With $Z_{\mathrm{gt}}$ the normalized ground-truth latents and $\epsilon\sim\mathcal N(0,\mathbf I)$,
\begin{equation}
Z_t=(1-t)Z_{\mathrm{gt}}
+\left[\sigma_{\min}+(1-\sigma_{\min})t\right]\epsilon,
\qquad
v^*=(1-\sigma_{\min})\epsilon-Z_{\mathrm{gt}},
\label{eq:flow-path}
\end{equation}
where $\sigma_{\min}$ is a small residual noise scale, $t=1$ corresponds to pure noise, and $t=0$ to material latents with residual noise $\sigma_{\min}\epsilon$. The loss is
\begin{equation}
\mathcal L_{\mathrm{FM}}=
\mathbb E\!\left[
\operatorname{MSE}\!\left(
v_\theta(Z_t,t;O,G),v^*\right)
\right],
\label{eq:flow-loss}
\end{equation}
averaged over all valid tokens and channels, including observed coordinates. The observation representation remains fixed and is not noised along the flow path; it serves as conditioning input to the generator. The generation stream is supervised against complete-material targets over the entire surface, learning to reconstruct visible appearance from the fixed observations and to complete unobserved regions.

\noindent\textbf{Inference.}
Given dataset-provided or predicted geometry and cameras, we encode the mesh into $G$ and $\mathcal S$ and the filtered observations into anchors $O$ (Sec.~\ref{sec:alignment}). Starting from $Z_1\sim\mathcal N(0,\mathbf I)$ over $\mathcal S$, we integrate the learned velocity field with Euler steps, then de-normalize and decode the final latents with $D_{\mathrm{mat}}$ into full-surface PBR attributes for export. Optional KV caching reuses the observation stream's per-layer keys and values across sampling steps; Table~\ref{tab:runtime} reports its runtime and memory effects.
Inference uses 25 Euler steps without classifier-free guidance on the time grid $t_j=3u_j/(1+2u_j)$, $u_j=1-j/25$, $j=0,\ldots,25$.

\begin{table}[!t]
  \caption{PBR texture generation with given geometry. Per-view metrics are averaged over 16 views; fg denotes the GT foreground. A.: albedo; Met./Rough.: metallic/roughness MAE.}
\label{tab:pbr-generation}
  \centering
  \scriptsize
  \setlength{\tabcolsep}{2.4pt}
  \resizebox{\textwidth}{!}{%
  \begin{tabular}{lcccccccccccc}
    \toprule
    Method
           & \multicolumn{6}{c}{Shaded} & \multicolumn{6}{c}{PBR} \\
    \cmidrule(lr){2-7}\cmidrule(lr){8-13}
           & PSNR $\uparrow$ & PSNR$_{\mathrm{fg}}$ $\uparrow$ & SSIM $\uparrow$ & LPIPS $\downarrow$ & CLIP-I $\uparrow$ & FD$_{\mathrm{DINOv3}}$ $\downarrow$
           & A.\ PSNR $\uparrow$ & A.\ PSNR$_{\mathrm{fg}}$ $\uparrow$ & A.\ SSIM $\uparrow$ & A.\ LPIPS $\downarrow$
           & Met.\ $\downarrow$ & Rough.\ $\downarrow$ \\
    \midrule
    \multicolumn{13}{l}{\splitname{DTC-Tex}} \\
    Step1X-3D & 23.12 & 14.21 & 0.914 & 0.083 & 0.851 & 33.30 & -- & -- & -- & -- & -- & -- \\
    MV-Adapter & 24.64 & 15.73 & 0.917 & 0.077 & 0.858 & 33.17 & -- & -- & -- & -- & -- & -- \\
    Ink3D & 25.82 & 17.08 & 0.921 & 0.076 & 0.859 & 35.01 & -- & -- & -- & -- & -- & -- \\
    UniTEX & 25.42 & 16.54 & 0.925 & 0.071 & 0.877 & 29.24 & -- & -- & -- & -- & -- & -- \\
    MVPainter & 26.44 & 17.59 & 0.924 & 0.077 & 0.856 & 33.40 & 24.17 & 15.30 & 0.911 & 0.089 & 0.121 & 0.119 \\
    TEXGen & 27.24 & 18.33 & 0.932 & 0.072 & 0.848 & 51.11 & -- & -- & -- & -- & -- & -- \\
    LumiTex & 28.29 & 19.38 & 0.939 & 0.062 & 0.892 & 24.42 & 26.30 & 17.39 & 0.928 & 0.069 & 0.352 & 0.098 \\
    Hunyuan3D 2.1 & 26.80 & 17.89 & 0.926 & 0.070 & 0.877 & 28.18 & 24.00 & 15.09 & 0.910 & 0.079 & 0.142 & 0.079 \\
    TRELLIS.2 & 28.60 & 19.73 & 0.935 & 0.062 & 0.880 & 32.73 & 25.78 & 16.90 & 0.921 & 0.070 & 0.063 & 0.059 \\
    Pixal3D$^\star$ & 29.19 & 20.47 & 0.950 & 0.050 & 0.908 & 20.35 & 26.39 & 17.60 & 0.937 & 0.057 & 0.051 & 0.076 \\
    \midrule
    \rowcolor{supraTableWash}
    Ours & 30.89 & 22.34 & 0.961 & \textbf{0.042} & 0.927 & 16.51 & 28.20 & 19.55 & 0.951 & \textbf{0.048} & 0.043 & 0.041 \\
    \rowcolor{supraTableWash}
    \textbf{Ours$^\ddagger$} & \textbf{32.08} & \textbf{23.67} & \textbf{0.962} & \textbf{0.042} & \textbf{0.928} & \textbf{16.31} & \textbf{29.68} & \textbf{21.14} & \textbf{0.955} & \textbf{0.048} & \textbf{0.033} & \textbf{0.038} \\
    \midrule
    \multicolumn{13}{l}{\splitname{Complex-Tex}} \\
    Step1X-3D & 23.61 & 13.97 & 0.924 & 0.066 & 0.848 & 30.93 & -- & -- & -- & -- & -- & -- \\
    MV-Adapter & 24.59 & 14.95 & 0.926 & 0.064 & 0.849 & 31.19 & -- & -- & -- & -- & -- & -- \\
    Ink3D & 26.15 & 16.69 & 0.935 & 0.060 & 0.864 & 30.86 & -- & -- & -- & -- & -- & -- \\
    UniTEX & 25.78 & 16.19 & 0.940 & 0.056 & 0.872 & 27.42 & -- & -- & -- & -- & -- & -- \\
    MVPainter & 25.72 & 16.19 & 0.933 & 0.067 & 0.846 & 33.87 & 23.05 & 13.51 & 0.922 & 0.079 & 0.228 & 0.188 \\
    TEXGen & 27.90 & 18.25 & 0.945 & 0.053 & 0.871 & 32.89 & -- & -- & -- & -- & -- & -- \\
    LumiTex & 27.77 & 18.21 & 0.947 & 0.051 & 0.887 & 23.29 & 25.26 & 15.68 & 0.938 & 0.060 & 0.256 & 0.181 \\
    Hunyuan3D 2.1 & 25.83 & 16.19 & 0.931 & 0.063 & 0.857 & 31.05 & 22.89 & 13.25 & 0.918 & 0.074 & 0.247 & 0.156 \\
    TRELLIS.2 & 27.97 & 18.41 & 0.945 & 0.051 & 0.876 & 27.42 & 25.77 & 16.17 & 0.935 & 0.058 & 0.131 & 0.111 \\
    Pixal3D$^\star$ & 28.84 & 19.44 & 0.950 & 0.046 & 0.899 & 19.50 & 26.62 & 17.25 & 0.935 & 0.056 & 0.145 & 0.122 \\
    \midrule
    \rowcolor{supraTableWash}
    Ours & 30.05 & 20.64 & 0.960 & 0.039 & 0.915 & 17.36 & 27.61 & 18.12 & 0.952 & \textbf{0.046} & \textbf{0.109} & 0.084 \\
    \rowcolor{supraTableWash}
    \textbf{Ours$^\ddagger$} & \textbf{30.66} & \textbf{21.29} & \textbf{0.961} & \textbf{0.038} & \textbf{0.921} & \textbf{16.08} & \textbf{27.74} & \textbf{18.27} & \textbf{0.953} & 0.047 & 0.115 & \textbf{0.076} \\
    \bottomrule
  \end{tabular}}
  \begin{minipage}{\linewidth}
    \raggedright\scriptsize
    $\ddagger$ A variant whose 2D material estimator is fine-tuned from Qwen-Image-2.1 instead of FireRed-Image-Edit-1.1.
  \end{minipage}
\end{table}

\section{Experiments}
\label{sec:experiments}
The supplement covers datasets, implementation, baseline configurations, evaluation protocols, and additional analyses.

\noindent\textbf{Datasets.}
Our 3D material generator is trained solely on publicly available 3D datasets. The training sources are summarized in Appendix~\ref{app:data-curation}. We evaluate on two datasets with no overlap with the training set: DTC-Tex, comprising 100 texture-rich product scans selected from the Digital Twin Catalog~\citep{dong2025dtc}, and Complex-Tex, comprising 110 Sketchfab~\citep{sketchfab} meshes with denser inscriptions, finer patterns, and more varied shapes.

\noindent\textbf{Implementation details.}
The 2D material estimator adapts FireRed-Image-Edit-1.1~\citep{firered2026imageedit} at $1024\times1024$ resolution. The 3D generator uses 60 Transformer blocks, with self-attention initialized from TRELLIS.2 and the remaining components initialized randomly; the pretrained material encoder and decoder remain frozen. We train the generator with a global batch size of 576 using AdamW on 64 GPUs. Inference uses 25 Euler steps without classifier-free guidance.

\begin{figure}[!t]
  \centering
  \includegraphics[width=\linewidth,page=1,trim=0bp 8.61bp 0bp 6.61bp,clip]{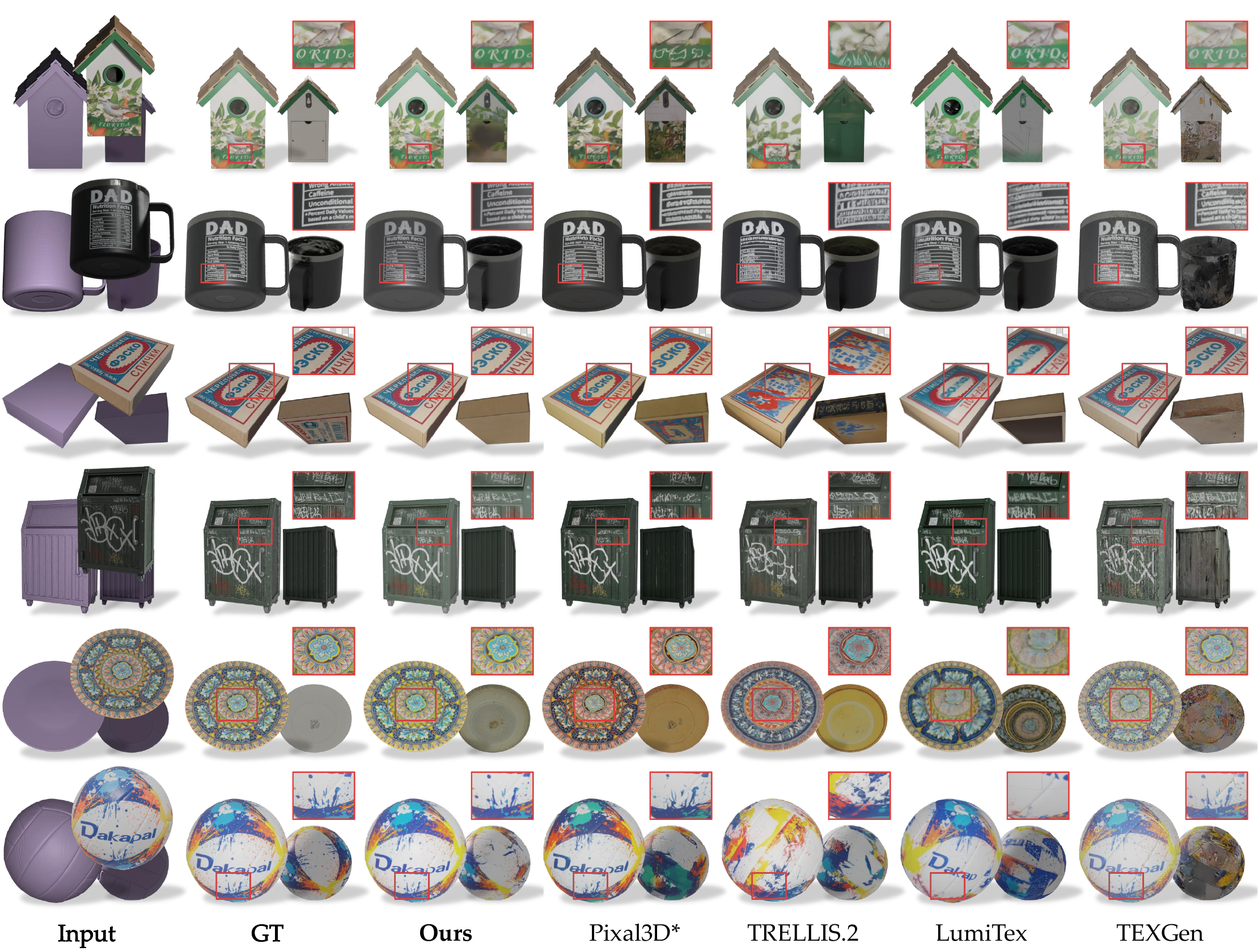}
  \caption{Qualitative comparison with given geometry. Enlarged crops compare reference-view details; novel views show the completed appearance.}
  \label{fig:pbr-qualitative}
  \vspace{10pt}
\end{figure}

\subsection{3D Generation with Given Geometry}
\label{sec:pbr-generation}

\noindent\textbf{Setting and baselines.}
Each method textures the ground-truth mesh from a single forward-lit reference image. We record the ground-truth lighting and camera poses and render each textured result under these same conditions at the reference view and 15 novel views, isolating material quality from geometry, pose, and illumination differences. Table~\ref{tab:pbr-generation} compares ten baselines spanning native 3D generation and dedicated texturing methods. PBR methods are additionally evaluated on albedo, metallic, and roughness.

\noindent\textbf{Results.}
Table~\ref{tab:pbr-generation} shows that \method{} leads across all shaded and PBR metrics on both datasets. Compared with feature-conditioned Pixal3D$^\star$, shaded PSNR improves by 1.7\,dB and 1.2\,dB, and albedo PSNR$_{\mathrm{fg}}$ by 2.0\,dB and 0.9\,dB on DTC-Tex and Complex-Tex, respectively. The gains over multi-view PBR methods are larger: LumiTex trails by more than 2\,dB in shaded PSNR and by over 1.5\,dB in albedo PSNR on both datasets. The albedo gains and lowest metallic/roughness MAE show that the improvements extend to material attributes. Figure~\ref{fig:pbr-qualitative} shows that \method{} better preserves the mug's fine lettering and the plate's intricate patterns than multi-view and feature-conditioned baselines, while novel-view renderings show coherent appearance on the completed surfaces.

\begin{table}[!t]
\centering
\caption{Reference/novel-view fidelity on the valid objects per method and metric in Table~\ref{tab:pbr-generation}. PSNR$_{\mathrm{fg}}$ uses the GT mask.}
\label{tab:view-split}
\supratablefont
\setlength{\tabcolsep}{3pt}
\begin{suprafulltable}{lrrrrrrrr}{16}
\toprule
& \multicolumn{4}{c}{Reference view (1)}
& \multicolumn{4}{c}{Novel views (15)} \\
\cmidrule(lr){2-5}\cmidrule(lr){6-9}
& \multicolumn{2}{c}{Shaded}
& \multicolumn{2}{c}{Albedo}
& \multicolumn{2}{c}{Shaded}
& \multicolumn{2}{c}{Albedo} \\
Method
& PSNR$_{\mathrm{fg}}\uparrow$ & LPIPS$\downarrow$
& PSNR$_{\mathrm{fg}}\uparrow$ & LPIPS$\downarrow$
& PSNR$_{\mathrm{fg}}\uparrow$ & LPIPS$\downarrow$
& PSNR$_{\mathrm{fg}}\uparrow$ & LPIPS$\downarrow$ \\
\midrule
\multicolumn{9}{l}{\splitname{DTC-Tex}} \\
Step1X-3D & 14.288 & 0.0861 & -- & -- & 14.207 & 0.0831 & -- & -- \\
MV-Adapter & 16.079 & 0.0806 & -- & -- & 15.706 & 0.0766 & -- & -- \\
Ink3D & 16.365 & 0.0795 & -- & -- & 17.128 & 0.0757 & -- & -- \\
UniTEX & 16.132 & 0.0711 & -- & -- & 16.572 & 0.0712 & -- & -- \\
MVPainter & 16.789 & 0.0815 & 15.530 & 0.0912 & 17.640 & 0.0770 & 15.286 & 0.0887 \\
TEXGen & 22.814 & 0.0380 & -- & -- & 18.029 & 0.0746 & -- & -- \\
LumiTex & 18.598 & 0.0594 & 18.083 & 0.0643 & 19.434 & 0.0620 & 17.342 & 0.0693 \\
Hunyuan3D 2.1 & 16.553 & 0.0696 & 15.206 & 0.0768 & 17.979 & 0.0700 & 15.079 & 0.0789 \\
TRELLIS.2 & 18.609 & 0.0660 & 17.093 & 0.0713 & 19.803 & 0.0621 & 16.885 & 0.0698 \\
Pixal3D$^\star$ & 21.837 & 0.0304 & 19.884 & 0.0363 & 20.377 & 0.0510 & 17.450 & 0.0585 \\
    \midrule
    \rowcolor{supraTableWash}
\textbf{Ours} & \textbf{25.372} & \textbf{0.0132} & \textbf{23.870} & \textbf{0.0156} & \textbf{22.134} & \textbf{0.0444} & \textbf{19.264} & \textbf{0.0500} \\
\midrule
\multicolumn{9}{l}{\splitname{Complex-Tex}} \\
Step1X-3D & 13.431 & 0.0737 & -- & -- & 14.003 & 0.0651 & -- & -- \\
MV-Adapter & 14.748 & 0.0708 & -- & -- & 14.963 & 0.0630 & -- & -- \\
Ink3D & 15.801 & 0.0639 & -- & -- & 16.752 & 0.0595 & -- & -- \\
UniTEX & 15.261 & 0.0593 & -- & -- & 16.249 & 0.0562 & -- & -- \\
MVPainter & 14.927 & 0.0769 & 13.430 & 0.0881 & 16.272 & 0.0662 & 13.512 & 0.0783 \\
TEXGen & 21.974 & 0.0266 & -- & -- & 18.005 & 0.0551 & -- & -- \\
LumiTex & 17.419 & 0.0516 & 15.722 & 0.0607 & 18.264 & 0.0514 & 15.677 & 0.0604 \\
Hunyuan3D 2.1 & 15.041 & 0.0694 & 13.145 & 0.0808 & 16.268 & 0.0624 & 13.262 & 0.0739 \\
TRELLIS.2 & 17.123 & 0.0561 & 15.545 & 0.0623 & 18.493 & 0.0503 & 16.209 & 0.0577 \\
Pixal3D$^\star$ & 20.062 & 0.0389 & 18.291 & 0.0453 & 19.402 & 0.0461 & 17.180 & 0.0566 \\
    \midrule
    \rowcolor{supraTableWash}
\textbf{Ours} & \textbf{22.927} & \textbf{0.0202} & \textbf{21.721} & \textbf{0.0254} & \textbf{20.488} & \textbf{0.0401} & \textbf{17.882} & \textbf{0.0474} \\
\bottomrule
\end{suprafulltable}
  \vspace{10pt}
\end{table}

\begin{table}[!t]
  \caption{End-to-end single-image-to-3D generation by dataset; column-wise best results are bold. Reconstruction uses four views except for reference-view CLIP-I; distribution metrics use 16-view rendered sets.}
  \label{tab:image-to-3d}
  \centering
  \supratablefont
  \setlength{\tabcolsep}{3pt}
  \begin{suprafulltable}{lccccccccc}{18}
    \toprule
    Method
           & \multicolumn{5}{c}{Reconstruction} & \multicolumn{4}{c}{Distribution} \\
    \cmidrule(lr){2-6}\cmidrule(lr){7-10}
           & PSNR $\uparrow$ & PSNR$_{\mathrm{fg}}$ $\uparrow$ & SSIM $\uparrow$ & LPIPS $\downarrow$ & CLIP-I $\uparrow$
           & FD$_{\mathrm{DINOv3}}$ $\downarrow$ & FD$_{\mathrm{CLIP}}$ $\downarrow$ & FID $\downarrow$ & CMMD $\downarrow$ \\
    \midrule
    \multicolumn{10}{l}{\splitname{DTC-Tex}} \\
    Step1X-3D & 19.70 & 12.00 & 0.894 & 0.115 & 0.844 & 33.4 & 19.6 & 33.5 & 14.40 \\
    Hunyuan3D 2.1 & 21.97 & 14.52 & 0.904 & 0.095 & 0.835 & 34.8 & 22.7 & 33.6 & 12.78 \\
    UniLat3D$^\dagger$ & 19.78 & 11.89 & 0.900 & 0.105 & 0.822 & 34.2 & 27.2 & 36.1 & 27.17 \\
    CUPID$^\dagger$ & 20.26 & 12.38 & 0.906 & 0.093 & 0.850 & 32.1 & 25.1 & 33.0 & 24.18 \\
    TRELLIS.2$^\dagger$ & 22.59 & 15.71 & 0.907 & 0.098 & 0.816 & 41.7 & 26.2 & 32.2 & 16.63 \\
    Pixal3D$^\star$ & 26.88 & 19.90 & 0.945 & 0.071 & 0.910 & 27.2 & 16.5 & 26.0 & 8.29 \\
    \midrule
    \rowcolor{supraTableWash}
    \textbf{Ours} & \textbf{28.65} & \textbf{20.79} & \textbf{0.967} & \textbf{0.025} & \textbf{0.971} & \textbf{21.1} & \textbf{9.9} & \textbf{23.0} & \textbf{4.97} \\
    \midrule
    \multicolumn{10}{l}{\splitname{Complex-Tex}} \\
    Step1X-3D & 20.66 & 12.36 & 0.900 & 0.099 & 0.882 & 26.3 & 17.9 & 31.4 & 8.73 \\
    Hunyuan3D 2.1 & 21.87 & 13.83 & 0.908 & 0.087 & 0.843 & 31.2 & 23.2 & 35.5 & 11.39 \\
    UniLat3D$^\dagger$ & 20.41 & 11.89 & 0.902 & 0.096 & 0.830 & 29.8 & 27.5 & 35.2 & 21.57 \\
    CUPID$^\dagger$ & 20.68 & 12.12 & 0.905 & 0.090 & 0.852 & 28.1 & 24.4 & 34.1 & 18.44 \\
    TRELLIS.2$^\dagger$ & 22.22 & 14.32 & 0.909 & 0.087 & 0.830 & 32.6 & 23.4 & 32.0 & 10.49 \\
    Pixal3D$^\star$ & 26.24 & 18.64 & 0.940 & 0.070 & 0.909 & 22.9 & 15.9 & 25.8 & 5.96 \\
    \midrule
    \rowcolor{supraTableWash}
    \textbf{Ours} & \textbf{28.73} & \textbf{20.24} & \textbf{0.963} & \textbf{0.029} & \textbf{0.968} & \textbf{19.4} & \textbf{9.3} & \textbf{23.3} & \textbf{4.00} \\
    \bottomrule
  \end{suprafulltable}
  \begin{minipage}{\linewidth}
    \raggedright\scriptsize
    $^\dagger$Similarity-aligned to the ground-truth mesh.
  \end{minipage}
\end{table}


Table~\ref{tab:view-split} separates the designated input view from the other 15 evaluation views, averaging novel-view scores per object before averaging over objects. For each method and metric, both splits use the same valid objects as Table~\ref{tab:pbr-generation}. Novel views include both previously observed and unobserved surfaces; this split measures viewpoint generalization, rather than error exclusively on unseen surface regions.

\method{} achieves the best mean in every reported metric on both view splits and datasets. Against the strongest baseline for shaded PSNR$_{\mathrm{fg}}$ in each split, reference-view gains are $2.56$\,dB on DTC-Tex and $0.95$\,dB on Complex-Tex (TEXGen), while novel-view gains are $1.76$ and $1.09$\,dB (Pixal3D$^\star$). The split supports improved reference-view fidelity together with improved novel-view appearance.

\subsection{End-to-End Single-Image-to-3D Generation}
\label{sec:image-to-3d}

\noindent\textbf{Setting and baselines.}
Each method generates both geometry and appearance end-to-end from a single reference image, without access to ground-truth geometry or cameras during generation.

We compare with TRELLIS.2, Pixal3D$^\star$, Hunyuan3D 2.1, Step1X-3D, UniLat3D, and CUPID~\citep{xiang2026native,li2026pixal3d,hunyuan3d2025hunyuan3d,li2025step1x3d,wu2025unilat3d,huang2025cupid}. \method{} and Pixal3D$^\star$ share geometry predicted by the latter, with separate appearance pipelines. Other baselines generate their own geometry. For scoring, outputs are brought into a common coordinate frame, with GT similarity alignment applied to methods marked $\dagger$ in Table~\ref{tab:image-to-3d}; this alignment plays no role in generation.
\par

\noindent\textbf{Results.}
Table~\ref{tab:image-to-3d} shows that \method{} leads across all $9$ metrics on both datasets. Relative to Pixal3D$^\star$, four-view PSNR improves by 1.77\,dB on DTC-Tex and 2.49\,dB on Complex-Tex, foreground PSNR by 0.89\,dB and 1.60\,dB, and LPIPS decreases by 64.8\% and 58.6\%, respectively. Higher reference-view CLIP-I and lower 16-view distribution distances further indicate stronger input consistency and multi-view appearance quality. Since Pixal3D$^\star$ and \method{} share predicted geometry, these results show that our appearance pipeline retains its advantage on predicted surfaces, complementing the given-geometry results in Table~\ref{tab:pbr-generation}.

\begin{figure}[!t]
  \centering
  \includegraphics[width=\linewidth,page=1,trim=0bp 15.97bp 0bp 10.98bp,clip]{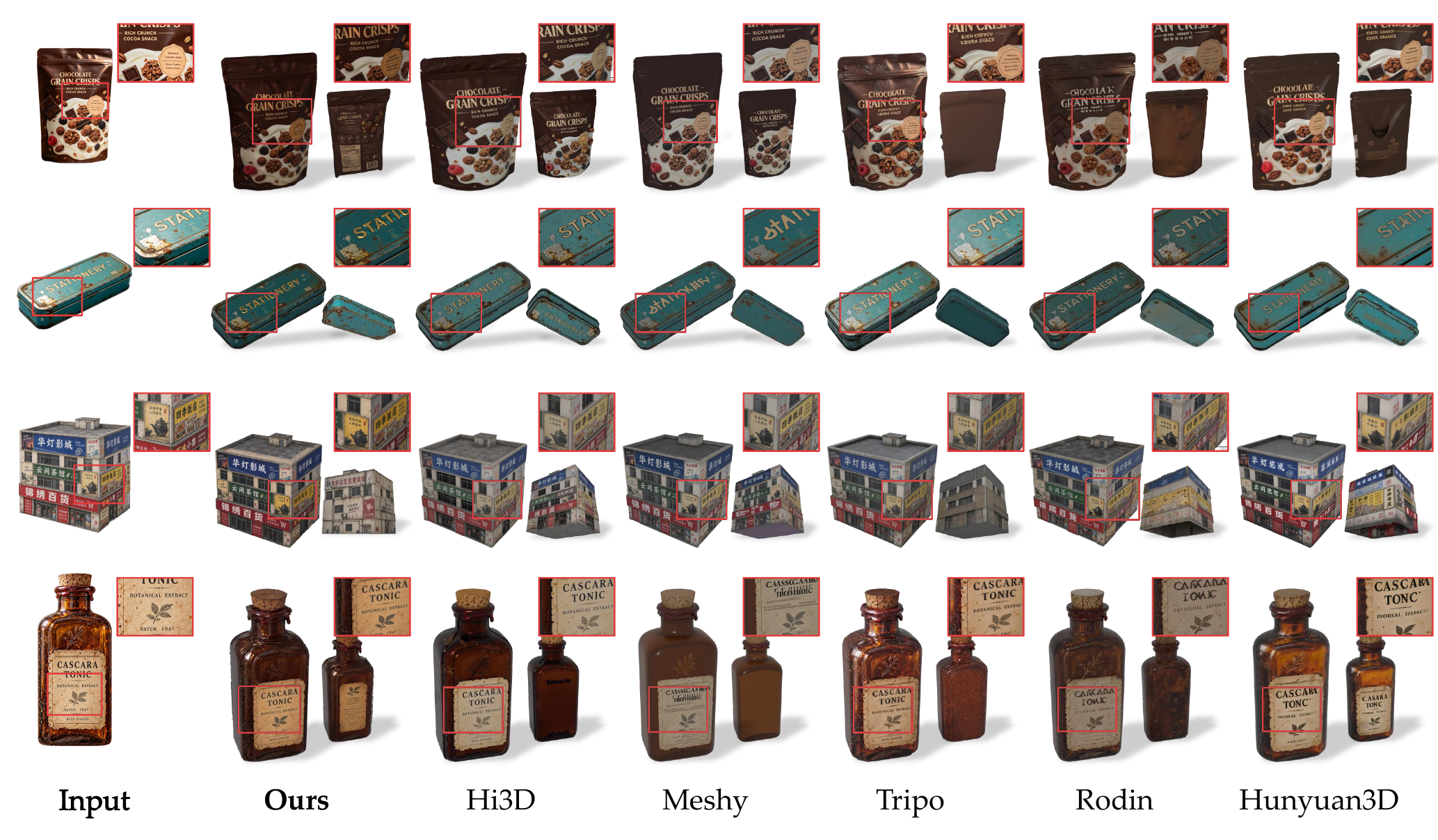}
  \caption{Comparison with commercial systems: each row shows the input and reference-view/back-view renderings from commercial systems and \method{}.}
  \label{fig:commercial}
\end{figure}

\subsection{Commercial Comparisons}
\label{sec:commercial}

We evaluate \method{} against five commercial systems on 45 real and synthetic images. We use two-view conditioning only for commercial comparisons, combining the reference image with a back view synthesized from it.

\begin{table}[!t]
  \centering
  \begin{minipage}[t]{0.46\linewidth}
    \vspace{0pt}
\caption{Comparison with commercial models. Preference reports our pairwise win rate (\%); VLM reports mean scores out of 10.}
  \label{tab:commercial}
  \centering
  \supratablefont
  \setlength{\tabcolsep}{2pt}
  \begin{suprafulltable}{lcccccc}{12}
    \toprule
    System & \multicolumn{3}{c}{Ours win rate (\%) $\uparrow$} & \multicolumn{3}{c}{VLM ($0$--$10$) $\uparrow$} \\
    \cmidrule(lr){2-4}\cmidrule(lr){5-7}
           & Align & PBR & Quality & Align & PBR & Quality \\
    \midrule
    HY3.1 & 84.7 & 85.1 & 84.8 & 6.77 & \textbf{8.67} & 8.32 \\
    Rodin 2.5 & 75.8 & 72.3 & 75.5 & 6.81 & 8.64 & \textbf{8.38} \\
    Tripo H3.1 & 71.6 & 62.9 & 67.8 & 7.71 & 8.53 & 8.28 \\
    Meshy 7.1 & 60.8 & 54.7 & 57.6 & 7.74 & 8.61 & 8.34 \\
    Hi3D 3.0 & 51.1 & 44.9 & 51.1 & \textbf{8.19} & \textbf{8.67} & 8.37 \\
    \midrule
    \rowcolor{supraTableWash}
    \textbf{Ours} & -- & -- & -- & 8.08 & \textbf{8.67} & 8.34 \\
    \bottomrule
  \end{suprafulltable}
  \end{minipage}\hfill
  \begin{minipage}[t]{0.52\linewidth}
    \vspace{0pt}
\caption{Observation ablations. Shad./Alb.: shaded/albedo PSNR$_{\mathrm{fg}}$; MR: mean metallic/roughness MAE.}
  \label{tab:ablations}
  \centering
  \supratablefont
  \setlength{\tabcolsep}{1.4pt}
  \begin{suprafulltable}{lcccccc}{12}
    \toprule
    Variant & \multicolumn{3}{c}{\splitname{DTC-Tex}} & \multicolumn{3}{c}{\splitname{Complex-Tex}} \\
    \cmidrule(lr){2-4}\cmidrule(lr){5-7}
            & Shad. $\uparrow$ & Alb. $\uparrow$ & MR $\downarrow$
            & Shad. $\uparrow$ & Alb. $\uparrow$ & MR $\downarrow$ \\
    \midrule
    Residual injection & 21.47 & 18.81 & 0.086 & 20.35 & 17.18 & 0.136 \\
    Joint-update attention & 22.32 & 19.40 & 0.043 & 20.63 & 17.96 & 0.097 \\
    \midrule
    Fixed MR ($M\!=\!0, R\!=\!0.5$) & 21.90 & 19.53 & 0.064 & 19.95 & 18.15 & 0.189 \\
    Dataset-mean MR & 22.08 & 19.53 & 0.232 & 20.07 & 18.13 & 0.258 \\
    \midrule
    \rowcolor{supraTableWash}
    Ours & 22.34 & 19.55 & 0.042 & 20.64 & 18.12 & 0.096 \\
    \midrule
    Oracle PBR (GT) & 24.65 & 22.17 & 0.011 & 21.76 & 20.63 & 0.030 \\
    \bottomrule
  \end{suprafulltable}
  \end{minipage}
\end{table}

Without ground-truth geometry or cameras, paired human preferences and VLM ratings assess input alignment, PBR plausibility, and visual quality. Trained solely on public data, \method{} receives \mbox{51.1\%--84.8\%} preference in human visual-quality comparisons and ties for the highest VLM PBR score of 8.67 (Table~\ref{tab:commercial}). Figure~\ref{fig:commercial} illustrates reference-detail preservation and coherent surface completion.
\par

\subsection{Ablation and Analysis}
\label{sec:ablations}
We evaluate the observation-injection architecture and material content under the given-geometry protocol of Sec.~\ref{sec:pbr-generation}. The architecture variants share training data, initialization protocol, optimization settings, training budget, and checkpoint-selection rule. All comparisons use the same geometry, cameras, valid observation support, and sampling seed.
\par

\noindent\textbf{How observations are used.}
Table~\ref{tab:ablations} compares three conditioning designs with the same predicted PBR observations. \emph{Residual injection} adds encoded observations through layerwise residuals. \emph{Joint-update attention} updates observation states in selected layers and all generation states in every layer. In contrast, \method{} keeps observations read-only and alternates unobserved-cell and full-surface updates (Sec.~\ref{sec:completion}). It improves shaded PSNR$_{\mathrm{fg}}$ over residual injection by 0.87\,dB on DTC-Tex and 0.29\,dB on Complex-Tex, and matches joint-update attention in shaded fidelity while improving albedo PSNR$_{\mathrm{fg}}$ by 0.15 and 0.16\,dB with lower MR MAE. Thus, the design preserves or improves reconstruction quality without spending computation on observation-state updates or full-surface queries at every block. The complete design also samples $1.20\times$ and $1.18\times$ faster than joint-update attention on the two datasets (Sec.~\ref{sec:runtime-memory}).

\noindent\textbf{Effectiveness of 2D material estimation.}
To test whether the FireRed-based 2D material estimator supplies useful physical material estimates, we replace its predicted metallic and roughness with fixed values ($M=0, R=0.5$) or dataset means, keeping predicted albedo and the 3D generator unchanged (Table~\ref{tab:ablations}). Relative to fixed and dataset-mean MR, predicted MR improves shaded PSNR$_{\mathrm{fg}}$ by 0.44 and 0.26\,dB on DTC-Tex, and 0.69 and 0.57\,dB on Complex-Tex, respectively, with lower MR MAE on both datasets. These gains show that the 2D estimates provide useful material information beyond albedo alone. Replacing all predicted PBR channels with ground truth (Oracle PBR) further improves shaded PSNR$_{\mathrm{fg}}$ by 2.31/1.12\,dB and albedo PSNR$_{\mathrm{fg}}$ by 2.62/2.51\,dB on DTC-Tex/Complex-Tex. The remaining gap indicates that improving 2D material estimation can further improve 3D texture quality without changing the completion model. Using Qwen-Image-2.1~\citep{qwen2026image21} (Ours$^\ddagger$ in Table~\ref{tab:pbr-generation}) narrows this oracle gap to 0.98/0.47\,dB in shaded PSNR$_{\mathrm{fg}}$ and 1.03/2.36\,dB in albedo PSNR$_{\mathrm{fg}}$. The same completion model therefore works with a different 2D prior, and a stronger prior improves 3D texture quality toward the oracle.

\begin{figure}[t]
  \centering
  \includegraphics[width=\linewidth]{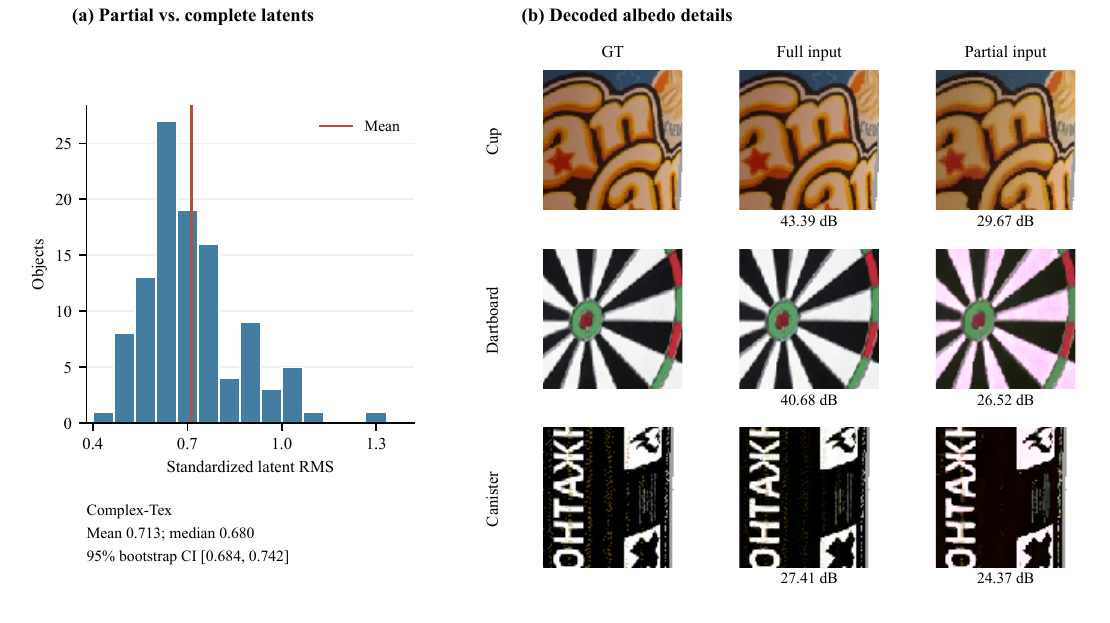}
  \caption{Native observation encoding with GT geometry and materials. (a) Partial-versus-complete latent differences across valid Complex-Tex objects. (b) Decoded albedo for three cases selected by latent-difference percentile, with shared GT-selected crops. PSNR covers all retained observed GT voxels, not only the displayed crops.}
  \label{fig:observation-analysis}
  \vspace{10pt}
\end{figure}

\phantomsection
\noindent\textbf{Partial and complete material encoding.}
\label{sec:observation-analysis}
Sharing the material encoder does not imply that a partial observation yields the corresponding slice of a complete-material latent. We isolate this distinction using GT geometry, cameras, and PBR values on Complex-Tex, bypassing both 2D prediction and flow generation. For each object, we encode complete materials and the filtered visible subset with the same frozen encoder using its posterior mean, then compare features at identical observed coordinates. We measure the RMS difference over coordinates and channels after dividing each channel by its fixed material-latent standard deviation, and average the resulting scores equally across objects. We retain objects that pass the coordinate and projection checks, excluding failed projection-IoU checks or incompatible geometry/material support. All retained objects have nonzero differences: the mean standardized RMS is 0.713 (95\% object-bootstrap interval 0.684--0.742), with median 0.680, while repeated encoding of identical inputs gives zero difference (Fig.~\ref{fig:observation-analysis}a). Thus, partial encoding changes representations even for unchanged input materials.

Latent distance alone does not establish material degradation. We decode the preselected cases nearest the 10th, 50th, and 90th latent-difference percentiles with the same frozen decoder. Each arm retains its encoder's spatial correspondence cache, so this tests material reconstruction on supplied geometry rather than autonomous recovery of surface support. Figure~\ref{fig:observation-analysis}b compares identical GT-selected crops; albedo PSNR is computed on the same retained GT surface voxels for both arms. Full/partial PSNR is 43.39/29.67, 40.68/26.52, and 27.41/24.37\,dB for the cup, dartboard, and canister, respectively. Partial reconstructions retain the main patterns but introduce additional detail and color errors, including a pink shift in the dartboard's white regions. The canister also shows appreciable full-input VAE error, which must be separated from the additional partial-input error. These three examples demonstrate decodable material content with imperfect fidelity; they are not a cohort-wide reconstruction estimate. Together, these results show that partial encoding supplies material evidence for refinement during full-surface generation. Appendix~\ref{app:observation-analysis} provides diagnostic settings.

\begin{table}[t]
\caption{Runtime and sampling memory. Speedup is relative to Joint-update attention. Peak mean/max summarize sampling allocations; stage sum adds separately timed pipeline stages, not continuous end-to-end latency.}
\label{tab:runtime}
\centering
\supratablefont
\begin{suprafulltable}{lrrrrr}{10}
\toprule
Configuration & Sampling (s)$\downarrow$ & Speedup & Stage sum (s)$\downarrow$ & Peak mean (GiB) & Peak max (GiB) \\
\midrule
\multicolumn{6}{l}{\splitname{DTC-Tex}} \\
Joint-update attention & 10.103 & $1.000\times$ & 26.840 & 12.586 & 13.757 \\
    \rowcolor{supraTableWash}
Ours & 8.440 & $1.197\times$ & 25.177 & 12.255 & 13.794 \\
    \rowcolor{supraTableWash}
Ours + KV cache & 7.909 & $1.277\times$ & 24.647 & 14.191 & 16.978 \\
\midrule
\multicolumn{6}{l}{\splitname{Complex-Tex}} \\
Joint-update attention & 9.970 & $1.000\times$ & 21.542 & 12.562 & 14.293 \\
    \rowcolor{supraTableWash}
Ours & 8.440 & $1.181\times$ & 20.012 & 12.243 & 14.699 \\
    \rowcolor{supraTableWash}
Ours + KV cache & 7.931 & $1.257\times$ & 19.503 & 14.049 & 17.340 \\
\bottomrule
\end{suprafulltable}
  \vspace{10pt}
\end{table}

\subsection{Runtime and Memory}
\label{sec:runtime-memory}

Table~\ref{tab:runtime} compares Joint-update attention with \method{} on all 210 given-geometry cases, using one H800 GPU per request. Both receive identical geometry and observation latents and use 25 Euler steps, seed~42, FP32 parameters, BF16 autocast, and FlashAttention~3 without classifier-free guidance. We report the mean of per-object median times from three runs after two warmups. These runtime measurements use the 10,000-step Joint-update attention checkpoint and the 30,000-step \method{} checkpoint, separately from the matched-training quality ablations in Table~\ref{tab:ablations}; this timing comparison does not isolate interleaving alone.

\method{} achieves sampling speedups of $1.20\times$ on DTC-Tex and $1.18\times$ on Complex-Tex. Observation KV caching adds $1.067\times$ and $1.064\times$ speedup, including cache construction, with identical latent outputs at every Euler step. Mean peak allocation rises by 1.94 and 1.81\,GiB, respectively. The interleaved schedule also reduces query-token evaluations by 17.54\% and 16.91\% versus querying every cell at each block; these are not total FLOP reductions.

Stage sums exclude loading and file I/O. Geometry and observation preparation are timed once; estimation, sampling, and decoding use warm measurements. Peak memory covers only the resident denoiser and sampling tensors.

\begin{figure}[t]
  \centering
  \includegraphics[width=\linewidth]{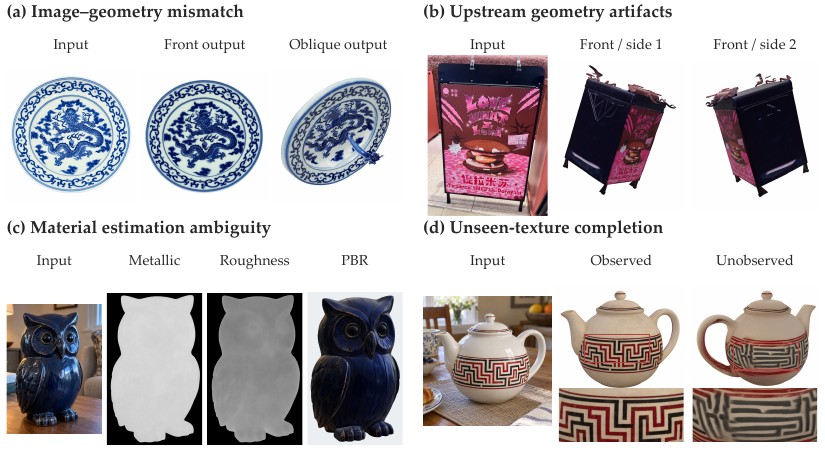}
  \caption{Limitations under imperfect inputs and incomplete observations.
(a) The front output exhibits local motif distortion around the dragon head; the oblique view reveals a spurious protrusion in the underlying geometry.
(b) Two front-oblique views show spurious protrusions in the upstream geometry that remain visible after texture refinement.
(c) The glossy painted-wood appearance is assigned a high metallic value. Without material ground truth, this example illustrates ambiguity in single-image material inference rather than a verified estimation error. Grayscale maps range from black (0) to white (1).
(d) Observed and unobserved base-color views show a loss of motif precision on the hidden side; boxes identify the enlarged details.
The front output in (a) and both outputs in (d) use unlit base color; the oblique output in (a) and both outputs in (b) use PBR rendering under the same neutral lighting.}
  \label{fig:limitations}

  \vspace{10pt}
\end{figure}

\section{Limitations}
\label{sec:limitations}

Figure~\ref{fig:limitations} shows how imperfect geometry, uncertain materials, and missing observations limit asset quality.

\noindent\textbf{Large image--geometry discrepancies.}
In end-to-end generation, \method{} uses geometry predicted by Pixal3D$^\star$. Large discrepancies in shape or pose between the reference image and the mesh can place projected material observations on incorrect surface locations. Correspondence filtering removes some unreliable observations, but cannot correct the underlying geometry, which remains fixed during material generation. Severe discrepancies can therefore prevent faithful texture alignment. Improving upstream geometry and camera alignment is therefore an important direction, but we do not evaluate corrected-geometry controls here.

\noindent\textbf{Hallucinated geometry on unseen surfaces.}
Upstream geometry generators such as Pixal3D$^\star$ can introduce implausible unseen structures even when the visible shape matches the reference. These structures may deviate from shapes learned by the material generator, hindering coherent texture completion. Since \method{} conditions on fixed geometry, it cannot remove these structures; their influence on texture generation can reduce final asset quality.

\noindent\textbf{Inconsistent completion across visibility boundaries.}
Completing unseen surfaces is particularly challenging when text or patterns extend beyond the visible region. For example, lettering that wraps around a can may be only partially visible in the input. The model may preserve the visible fragment yet fail to continue its strokes or layout coherently onto the hidden side, producing inconsistent text or patterns across the visibility boundary.

\noindent\textbf{Errors in 2D material estimation.}
The 2D material estimator can misinterpret material properties, for example by assigning high metallic values to a non-metallic surface. These estimates condition the 3D generator, which may retain the error in the completed asset and produce implausible reflections under rendering or relighting. Accurate transfer of reference details thus does not guarantee correct material decomposition.

\noindent\textbf{Failure cases.}
Figure~\ref{fig:limitations} illustrates these limitations. Reference fidelity depends on usable image--surface correspondence and reliable material observations, while fine detail on unseen surfaces must be generated. The plate shows a printed motif reconstructed as protruding geometry, and the sign retains spurious structures after texture refinement. The owl separates recognizable appearance details from an ambiguous reflectance estimate; the teapot separates sharp observed patterns from less precise unseen completion. These are qualitative cases, without corrected-geometry controls or measured unseen-surface ground truth. A back view may help the last case; we report no paired two-view result.

\section{Conclusion}
\label{sec:conclusion}

We presented \method{}, a framework for single-image 3D generation with super-aligned textures through observation-guided material completion. By separating visible-material estimation and surface alignment from 3D generation and using native observation latents as fixed, read-only anchors, \method{} preserves fine reference details while generating coherent, relightable materials across the full surface. Trained solely on public data, it achieves state-of-the-art results among the evaluated methods in geometry-conditioned texturing and end-to-end asset generation. With synthetic back-view conditioning, it is also competitive with commercial systems in human evaluations on the mixed real-and-synthetic comparison set. These results demonstrate the value of spatially aligned, directly usable material observations for reconciling reference fidelity with full-surface generation.

\clearpage
\bibliography{supratex}
\bibliographystyle{supratex}

\clearpage
\appendix
\input{appendix.tex}

\end{document}

%% file: math_commands.tex
\usepackage{amsmath,amsfonts,bm}

\def\eqref#1{equation~\ref{#1}}

\def\1{\bm{1}}

\DeclareMathAlphabet{\mathsfit}{\encodingdefault}{\sfdefault}{m}{sl}
\SetMathAlphabet{\mathsfit}{bold}{\encodingdefault}{\sfdefault}{bx}{n}



%% file: commands.tex
\newcommand{\methodname}{SUPRA-Tex}
\newcommand{\method}{\methodname}

\newcommand{\projectpageurl}{https://supra-3d.github.io/supratex}

%% file: authors.tex
\hypersetup{
  colorlinks=true,pdfborder={0 0 0},
  pdftitle={SUPRA-TEX: 3D Generation with Super-Aligned Texture},
  pdfauthor={Chen Yang, Jiemin Fang, Fanpeng Meng, Zihao Huang, Jichen Pang, Guanjun Wu, Zhikuan Bao, Qi Tian},
  pdfsubject={3D generation with super-aligned textures},
  pdfkeywords={3D generation, texture generation, PBR materials}
}

\author{%
  \hypersetup{urlcolor=black}%
  \makebox[60pt][l]{\textbf{\href{https://chensjtu.github.io/}{Chen Yang}}\textsuperscript{1*}}%
  \makebox[72pt][l]{\textbf{\href{https://jaminfong.cn/}{Jiemin Fang}}\textsuperscript{1*\textdagger\supracorresponding}}%
  \makebox[72pt][l]{\textbf{\href{https://mfp0610.github.io/}{Fanpeng Meng}}\textsuperscript{1*}}%
  \makebox[72pt][l]{\textbf{\href{https://inso-13.github.io/}{Zihao Huang}}\textsuperscript{1}}\\[2pt]
  \makebox[60pt][l]{\textbf{Jichen Pang}\textsuperscript{2}}%
  \makebox[72pt][l]{\textbf{\href{https://wuguanjun0627.github.io/}{Guanjun Wu}}\textsuperscript{3}}%
  \makebox[72pt][l]{\textbf{Zhikuan Bao}\textsuperscript{1}}%
  \makebox[72pt][l]{\textbf{\href{https://www.qitian1987.com/index.html}{Qi Tian}}\textsuperscript{1\supracorresponding}}\\[8pt]
  \textsuperscript{1}Huawei Inc.\quad
  \textsuperscript{2}Shanghai Jiao Tong University\quad
  \textsuperscript{3}Huazhong University of Science and Technology\\[6pt]
  \textsuperscript{*}Equal contribution\quad
  \textsuperscript{\textdagger}Project lead\quad
  \textsuperscript{\supracorresponding}Corresponding authors\\[7pt]
}

%% file: abstract.tex
Single-image 3D generation aims to dress a given 3D mesh with relightable materials while faithfully preserving the visual details of a reference image. Existing methods either introduce artifacts when fusing inconsistent multi-view predictions or struggle to preserve fine details through implicit image-feature conditioning. To resolve these challenges, we introduce \suprahighlight{SUPRA-Tex}, built on the principle of \emph{decoupling visible-material estimation, surface alignment, and observation-guided 3D completion}. This division of labor allows the 3D generator, trained on limited 3D data, to focus its capacity on preserving visible evidence and completing unobserved surfaces. Specifically, we first train an MLLM-based image editing model to predict lighting-separated PBR maps in the reference view. We then project these maps onto the visible surface and encode them with the pretrained material VAE encoder into \emph{native observation latents} that share the generation target's latent space. An \emph{interleaved flow transformer} holds these latents as fixed, read-only anchors while alternating unobserved-region and full-surface updates to generate complete material latents. This design achieves \emph{super-alignment} with the reference, delivering \emph{What You See Is What You Get} fidelity on visible regions and coherent completion on unobserved surfaces. \underline{Trained solely on public datasets}, SUPRA-Tex achieves SOTA performance in both geometry-conditioned texturing and end-to-end asset generation, matching or even exceeding leading commercial systems on in-the-wild images. Project page: \href{\projectpageurl}{\texttt{supra-3d.github.io/supratex}}.

%% file: appendix.tex
\section*{Supplementary Material}
\raggedbottom

This supplement provides method and training details, evaluation protocols, additional results, and extended discussion. The guide below links to each appendix and selected subsections; page numbers indicate where each appendix begins.

\begin{itemize}
\setlength{\itemsep}{6pt}
\setlength{\parskip}{0pt}
\item \hyperref[app:method-details]{\textbf{Appendix~\ref*{app:method-details}: Additional Method Details}} (p.~\pageref{app:method-details}).
Geometry and camera projection, visible-surface observations, and boundary filtering; the \hyperref[app:transformer-details]{interleaved flow transformer}, including token dimensions and its read-only update schedule. The training objective and inference procedure are given in Sec.~\ref{sec:training-inference}.

\item \hyperref[app:training-details]{\textbf{Appendix~\ref*{app:training-details}: Training Data and Optimization}} (p.~\pageref{app:training-details}).
\hyperref[app:data-curation]{Training data}, the \hyperref[app:material-estimator-training]{2D material estimator $F_{\mathrm{2D}}$}, and \hyperref[app:generator-training]{3D generator training}, including observation construction, optimization, and checkpoint selection.

\item \hyperref[app:evaluation-details]{\textbf{Appendix~\ref*{app:evaluation-details}: Evaluation Protocols}} (p.~\pageref{app:evaluation-details}).
Evaluation datasets, \hyperref[app:rendering-metrics]{rendering cameras, alignment, and metrics}, baseline configurations, and \hyperref[app:commercial-protocol]{commercial comparisons}. 

\item \hyperref[app:quantitative]{\textbf{Appendix~\ref*{app:quantitative}: 2D Material Evaluation}} (p.~\pageref{app:quantitative}).
\hyperref[app:2d-material]{2D material prediction accuracy} and \hyperref[app:material-estimator-runtime]{$F_{\mathrm{2D}}$ runtime measurements}. Reference-view versus novel-view fidelity and 3D sampling efficiency are reported in Secs.~\ref{sec:pbr-generation} and~\ref{sec:runtime-memory}.

\item \hyperref[app:ablations]{\textbf{Appendix~\ref*{app:ablations}: Ablation and Observation Analyses}} (p.~\pageref{app:ablations}).
Ablation settings, the \hyperref[app:observation-analysis]{partial-versus-complete encoding diagnostic protocol}, \hyperref[app:zero-filling]{observed-support encoding versus zero filling}, and \hyperref[app:native-pbr-learning]{early learning with native PBR latents}.

\item \hyperref[app:extensions]{\textbf{Appendix~\ref*{app:extensions}: Additional Qualitative Comparisons}} (p.~\pageref{app:extensions}).
Additional visual comparisons for given-geometry texturing, end-to-end image-to-3D generation, and commercial systems, extending the main-paper examples.

\item \hyperref[app:editing-limitations]{\textbf{Appendix~\ref*{app:editing-limitations}: Additional Views and Texture Editing}} (p.~\pageref{app:editing-limitations}).
\hyperref[app:texture-editing]{Synthetic back-view conditions and image-based texture editing}. Limitations and failure cases are discussed in Sec.~\ref{sec:limitations}.

\item \hyperref[app:notation]{\textbf{Appendix~\ref*{app:notation}: Notation Reference}} (p.~\pageref{app:notation}).
A consolidated symbol table with dimensions and definitions for the physical material channels, surface and latent coordinates, observation metadata, and generator states.
\end{itemize}

\clearpage

\section{Additional Method Details}
\label{app:method-details}
\suppressfloats[t]

\subsection{Geometry and Surface Observations}

\noindent\textbf{Geometry foundation and camera projection.}
Our framework operates on an underlying mesh $\mathcal M$ and camera projection $\pi$, provided as ground-truth inputs or estimated from the reference image. For end-to-end inference, we follow the default MoGe-2-based camera-estimation pipeline of Pixal3D$^\star$, whose geometry module generates the mesh. For the 3D position $\mathbf x_s$ of latent coordinate $s$, the model samples image features through camera projection:
\begin{equation}
 f_s=\operatorname{Sample}\!\left(\Phi(I),\pi(\mathbf x_s)\right),
\label{eq:projection}
\end{equation}
where $\Phi(I)$ denotes a 2D feature map. In the geometry refinement stages, Pixal3D$^\star$ projects both the original and spatially upsampled image features and concatenates the sampled features. Combined with global image conditioning, these features align the predicted mesh $\mathcal M$ and camera projection $\pi$ with the reference view, serving as the geometric input for our material generation.

\noindent\textbf{Surface observations and latent coordinates.}
The full surface defines the active latent-coordinate set $\mathcal S$. Valid visible surface voxels form the partial PBR input $\mathcal P_o$, which is encoded before any latent-level conditioning is constructed. The resulting coordinates $\mathcal S_o$ identify cells with observations, but a cell in $\mathcal S_o$ can also contain unseen surface voxels; $\mathcal U=\mathcal S\setminus\mathcal S_o$ therefore denotes cells without any observation, not all unseen surface. Coverage $\rho_s$ is the observed fraction of the surface voxels actually present in a cell, and repeated image samples of one voxel do not increase it. Boundary distance $d_s$ uses the six-neighbor graph of active latent coordinates and is $-1$ if no unobserved cell is reachable.

\noindent\textbf{Inference-time boundary filtering.}
At inference, we additionally apply foreground-mask erosion and local depth-discontinuity filtering to reduce texture contamination from unreliable correspondences. Foreground erosion removes mixed pixels along the silhouette, and depth-discontinuity filtering rejects correspondences whose neighboring visible depths differ by more than a threshold, where a small projection error would move the observation onto a different surface. All material channels share the retained observation support, from which local material encoding, coverage, and boundary distance are computed.

\subsection{Interleaved Flow Transformer}
\label{app:transformer-details}

\noindent\textbf{3D generator architecture.}
Material and geometry latents have 32 channels each on a sparse latent grid of resolution $64^3$. The generator has 60 Transformer blocks with hidden dimension 1536 and 12 attention heads, 30 of each update type; its self-attention modules are initialized from TRELLIS.2 and the remaining components randomly.

\noindent\textbf{Two states at an observed coordinate.}
At every $s\in\mathcal S_o$, the read-only observation token $O_s$ represents a partial material observation, while the generation token $X_s$ predicts the complete material of the cell. Geometry and material latents use fixed per-channel normalization; observation latents are mapped to the material-latent scale before the shared input projection. Projected spatial metadata and the learned role bias $b_o\in\mathbb R^{1536}$, initialized to zero, are added only to observation tokens. Table~\ref{tab:supp-representation} summarizes the dimensions.

\begin{table}[!htbp]
\centering
\caption{Representation dimensions. $N=|\mathcal S|$ and $M=|\mathcal S_o|$ vary across objects.}
\label{tab:supp-representation}
\small
\begin{tabular}{lll}
\toprule
Quantity & Dimensions & Meaning \\
\midrule
Geometry latents $G$ & $N\times32$ & Geometry conditions on the full surface \\
Noisy material latents $Z_t$ & $N\times32$ & Full-surface material state at time $t$ \\
Native observation latents $\{z_s^o\}_{s\in\mathcal S_o}$ & $M\times32$ & Encoded partial PBR materials \\
Observation metadata & $M\times2$ & Coverage and boundary distance \\
Generation stream $X$ & $N\times1536$ & Hidden generation states \\
Observation stream $O$ & $M\times1536$ & Read-only observation states \\
Output velocity & $N\times32$ & Updates to all material latents \\
\bottomrule
\end{tabular}
\end{table}

\begin{figure}[!t]
  \centering
  \includegraphics[width=\linewidth]{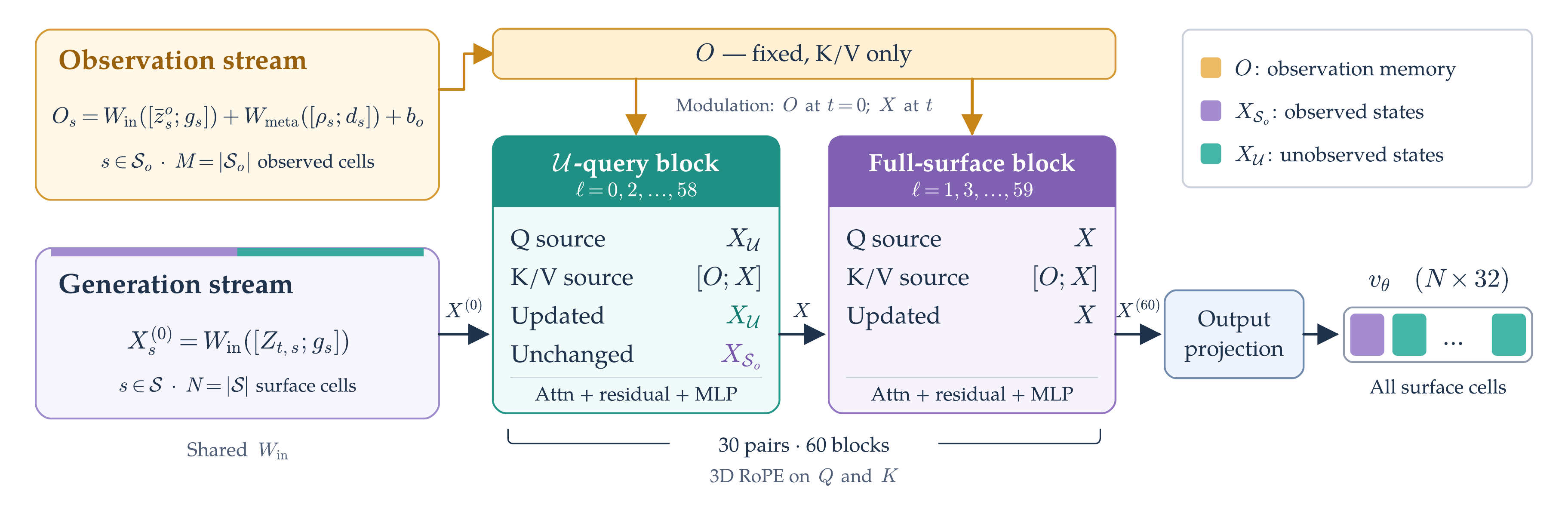}
  \caption{\textbf{Interleaved flow transformer.} One forward pass through 60 distinct blocks arranged in 30 alternating pairs. The fixed observation stream $O$ (orange) supplies keys and values without queries or state updates. The generation stream $X$ spans the full active latent-cell set $\mathcal S$, with states at observed cells $\mathcal S_o$ (purple) and cells without any observation $\mathcal U=\mathcal S\setminus\mathcal S_o$ (teal); the subscript $o$ denotes observed. An observed cell may still contain unseen surface voxels. Even-indexed blocks query and update $X_{\mathcal U}$; odd-indexed blocks query and update all of $X$. Both block types read keys and values from all of $O$ and $X$, and the output projection predicts velocities over all $N=|\mathcal S|$ cells. Source labels denote tokens before normalization, time modulation, and attention projections. At inference, 25 Euler sampling steps each invoke a full transformer forward pass.}
  \label{fig:interleaved-flow}
\end{figure}

\noindent\textbf{Attention and update schedule.}
Figure~\ref{fig:interleaved-flow} illustrates the two streams and their alternating update schedule. Each of the 60 Transformer blocks reads keys and values from both streams. Even-indexed blocks use queries from $X_{\mathcal U}$, odd-indexed blocks from all of $X$; generation states in $\mathcal U$ are therefore updated in 60 blocks and those in $\mathcal S_o$ in 30. Observation tokens pass through each block's normalization and adaptive modulation with the clean-time indicator $t=0$, produce no queries, and receive no residual or feed-forward updates. Queries from $X$, and keys from both $O$ and $X$, carry rotary positional encoding~\citep{su2024roformer} extended to 3D latent coordinates; values are not rotated. These counts describe hidden-state updates within one forward pass and do not restrict which material latents evolve during sampling: the output velocity and Euler updates cover all of $\mathcal S$.

\Needspace{7\baselineskip}
\section{Training Data and Optimization}
\label{app:training-details}

\subsection{Training Data}
\label{app:data-curation}

We train our models using publicly available 3D datasets. The 3D generator uses assets from TexVerse~\citep{zhang2025texverse}, Objaverse-XL~\citep{deitke2023objaversexl}, ABO~\citep{collins2022abo}, HSSD~\citep{khanna2024hssd}, and open synthetic 3D collections, together with human scans from 4D-Dress~\citep{wang20244ddress}, Human4DiT~\citep{shao2024human4dit}, THuman2.1~\citep{yu2021function4d}, and X-Humans~\citep{shen2023xavatar}. These assets provide geometry and material supervision for observation-guided full-surface completion. The 2D material estimator is trained on paired RGB images, material maps, and normal maps rendered from Objaverse assets. All datasets are used in accordance with their respective licenses for non-commercial academic research.

\subsection{Training \texorpdfstring{$F_{\mathrm{2D}}$}{F2D}}
\label{app:material-estimator-training}

We initialize $F_{\mathrm{2D}}$ from FireRed-Image-Edit-1.1 adapted to individual PBR and normal prediction tasks, and jointly fine-tune it on aligned RGB images and ground-truth material and normal maps rendered from publicly available 3D assets. Both the task-specific adaptation and joint fine-tuning use these rendered data. Modality-axis attention is inserted at Transformer blocks 29, 39, 49, and 59, with rank 256 and four attention heads. We optimize these adapters and backbone blocks 29--49 using a latent flow-matching objective and AdamW~\citep{loshchilov2019adamw}, with learning rates of $2\times10^{-4}$ and $5\times10^{-7}$, respectively. The VAE and text encoder remain frozen.

By folding the batch dimension, the model processes albedo, metallic-roughness, and an auxiliary normal stream in parallel, inserting lightweight cross-attention adapters at specific transformer layers to exchange features across modality axes at identical spatial coordinates. We train $F_{\mathrm{2D}}$ with a flow-matching objective anchored at the fixed maximum-noise endpoint $\sigma = 1.0$ using an outlier-masked MSE loss. Elementwise latent-flow residuals with absolute magnitude greater than 50 contribute zero loss but remain in the averaging denominator. Losses are averaged equally across the albedo, metallic-roughness, and normal streams. At inference, we use a single prediction at this endpoint to estimate the clean material latents, which the image VAE decodes into $1024\times 1024$ PBR maps.

\subsection{3D Generator Training}
\label{app:generator-training}

\noindent\textbf{Preprocessing for training.}
We build training observations from ground-truth PBR materials of public 3D assets using the visible-surface observation encoding pipeline. Foreground validity and first-hit visibility determine the observed surface voxels supplied to the sparse material encoder. We follow Pixal3D~\citep{li2026pixal3d} for asset preprocessing and reference-view sampling. During training, we drop the entire observation condition with probability $0.03$ while retaining the geometry condition.

\noindent\textbf{Optimization.}
We use random seed~42 throughout our training and evaluation pipelines. The material encoder and decoder are frozen. Generator self-attention is initialized from TRELLIS.2 and the remaining components randomly. We use the flow-matching objective in Eq.~\eqref{eq:flow-loss}.
We optimize the 3D generator for 30,000 steps with a global batch size of 576 using AdamW, with $\beta_1=0.9$, $\beta_2=0.95$, weight decay $0.01$, and $\epsilon=10^{-8}$. The initial learning-rate schedule uses 150 steps of linear warmup to $10^{-4}$, followed by cosine decay with a 20,000-step horizon and a minimum learning rate of $10^{-5}$. At step 15,000, we extend training with a cosine continuation from approximately $2.337\times10^{-5}$ to $10^{-5}$ at step 30,000, without additional warmup. We use BF16 automatic mixed precision and adaptive gradient clipping with a maximum norm of $1.0$ and a 95th-percentile threshold. The final training stage uses 64 GPUs and FlashAttention-3 acceleration. Evaluation uses the raw generator weights at step 30,000.

\noindent\textbf{Training-time sampling.}
We use $\sigma_{\min}=10^{-5}$ and sample training times as $t=\operatorname{sigmoid}(\xi)$ with $\xi\sim\mathcal N(1,1)$. Observation dropout is applied as described in preprocessing above.

\section{Evaluation Protocols}
\label{app:evaluation-details}

\subsection{Evaluation Datasets}

DTC-Tex comprises 100 objects from the Digital Twin Catalog~\citep{dong2025dtc}.
We score the $2{,}399$ catalog assets for surface-texture complexity from six views with Qwen3.8-27B, discard near-duplicates and broken meshes, and retain the 100 highest-scoring objects.
Each object is evaluated from a front-view reference, together with its ground-truth mesh, PBR maps, and a 16-view camera rig.
Typical cases are household products with printed labels or decorative motifs, such as canned goods, decorated ceramics, and books.

\begin{figure}[!t]
  \centering
  \includegraphics[width=\linewidth]{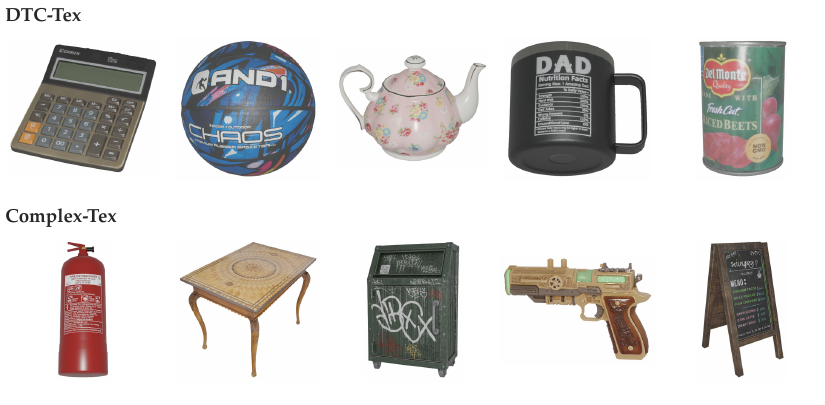}
  \caption{\textbf{Examples from DTC-Tex and Complex-Tex.} Ten selected assets from DTC-Tex (top) and Complex-Tex (bottom), shown in their designated reference views. Examples include dense text, fine patterns, and reflective surfaces.}
  \label{fig:dataset-examples}
\end{figure}

Complex-Tex complements this set with 110 Sketchfab~\citep{sketchfab} meshes that carry denser inscriptions and finer patterns on more varied shapes.
This subset includes branded packaging and vehicles covered by logos and fine print, as well as buildings and characters with more irregular shapes.
A method receives only the designated reference image, or that image together with the mesh when geometry is given.
Figure~\ref{fig:dataset-examples} shows examples from both datasets.

Both DTC-Tex and Complex-Tex underwent strict duplicate checks during data collection to ensure no overlap with the training assets. We will release both evaluation benchmarks.

\subsection{Rendering and Metrics}
\label{app:rendering-metrics}

\noindent\textbf{Protocol.}
Both settings evaluate the PBR asset each method exports, rendered under a shared protocol rather than compared in UV space, because latent decoding or remeshing changes the face layout. Predicted and ground-truth assets are rendered with the same cameras and forward lighting at $1024\times1024$ and downsampled to $512\times512$ with Lanczos interpolation; methods whose output lives in another coordinate frame are brought into the evaluation frame by their own transform or, where marked $\dagger$, by a similarity transform to the ground-truth mesh. Per-view metrics are PSNR, SSIM~\citep{wang2004ssim}, and VGG-LPIPS~\citep{zhang2018lpips} over the full image with a black background, so they penalize errors in silhouette as well as appearance, and CLIP-I, the cosine similarity of CLIP ViT-L/14 features~\citep{radford2021clip} between corresponding views. We additionally report PSNR$_{\mathrm{fg}}$ on both shaded RGB and albedo, the same PSNR restricted to the ground-truth object mask, so that large black background regions do not dominate the score. Distribution metrics compare multi-view image sets within each dataset: FD$_{\mathrm{DINOv3}}$ uses 1024-dimensional DINOv3 ViT-L/16 CLS features~\citep{simeoni2025dinov3}. In end-to-end evaluation, FID~\citep{heusel2017fid} uses Inception-v3 features, while FD$_{\mathrm{CLIP}}$ and CMMD~\citep{jayasumana2024cmmd} provide additional distribution measures. For methods that output PBR, we report mean absolute error (MAE) of metallic and roughness over the intersection of the predicted and ground-truth object masks. Per-view metrics are averaged over views per object, then over valid objects. We report results separately on DTC-Tex and Complex-Tex. In the given-geometry setting, the mesh and camera are provided as ground truth; end-to-end generation predicts geometry, aligned to GT by a similarity transform only for scoring.

\noindent\textbf{Rendering cameras and alignment.}
For given-geometry evaluation, we replay the 16 camera poses and front-light settings stored in each ground-truth asset's \texttt{transforms.json}; the designated input camera defines the reference view, and the other 15 cameras define novel views. Ground-truth and predicted assets use identical cameras and lighting, rendered with Blender~3.0.1 and Filmic color management. Where geometric alignment is required, we use iterative closest point (ICP) and silhouette-based refinement to estimate a similarity transform to the ground-truth geometry before rendering. This alignment is used only for evaluation.

\noindent\textbf{DINOv3 Fréchet distance for given-geometry texturing.}
For Table~\ref{tab:pbr-generation}, we extract the 1024-dimensional final-layer CLS token of DINOv3 ViT-L/16 without additional $\ell_2$ normalization. We alpha-composite each shaded rendering onto black, resize its shorter side to 256 pixels using bilinear interpolation, center-crop to $224\times224$, and apply ImageNet mean and standard-deviation normalization. These features are computed in FP32 with TF32 disabled using Transformers 4.57.3. We compute the Fréchet distance between the feature means and unbiased covariances of the generated and corresponding ground-truth image sets, using float64 statistics. This is one set-level score per method and dataset, rather than an average of object-level distances.

\subsection{Open-Source Baselines}

\noindent\textbf{Given-geometry comparison.}
Each object provides a ground-truth mesh and a forward-lit reference image from a specified view; every method textures the ground-truth mesh from this image through its own interface, and we render 16 views comprising the reference view and 15 novel views. We compare against the texturing stages of the native 3D generators TRELLIS.2 and Pixal3D$^\star$ and of the pipelines Hunyuan3D 2.1 and Step1X-3D~\citep{xiang2026native,li2026pixal3d,hunyuan3d2025hunyuan3d,li2025step1x3d}, and against the dedicated texturing methods LumiTex, MVPainter, MV-Adapter, TEXGen, UniTEX, and Ink3D~\citep{bao2025lumitex,shao2025mvpainter,huang2024mvadapter,yu2024texgen,liang2025unitex,han2026ink3d}. TRELLIS.2, Pixal3D$^\star$, Hunyuan3D 2.1, LumiTex, and MVPainter output PBR materials and are evaluated on shaded renderings, albedo, metallic, and roughness; the remaining methods output RGB textures and are evaluated on shaded renderings only, with unavailable channels marked ``--''.
We use the official implementations and released checkpoints for all baselines, providing the reference camera to every method that supports camera conditioning.

\noindent\textbf{End-to-end comparison.}
When geometry is also predicted, reference fidelity depends on geometry reconstruction, surface correspondence, and material generation together. This is the setting the observation interface is most exposed to: the anchors now land on a predicted mesh through a predicted camera, so any geometry or correspondence error propagates into the materials. Each method takes the reference image alone. Pixal3D$^\star$ and \method{} share the same predicted geometry and apply their respective appearance-generation pipelines; the other baselines use their own geometry-generation pipelines. \method{} textures this predicted mesh with the pipeline of Sec.~\ref{sec:training-inference}. We compare with TRELLIS.2, Pixal3D$^\star$, Hunyuan3D 2.1, Step1X-3D, UniLat3D, and CUPID~\citep{xiang2026native,li2026pixal3d,hunyuan3d2025hunyuan3d,li2025step1x3d,wu2025unilat3d,huang2025cupid}, covering native latent generation, staged geometry-then-texture pipelines, and joint object--pose modeling. Since output material attributes differ across these systems, we compare complete assets through rendered appearance only. For scoring, outputs are brought into a common coordinate frame using their coordinate transforms or, for methods marked $\dagger$, similarity alignment to the ground truth; this alignment plays no role in generation.

Reconstruction metrics take the reference view as anchor and compare four rendered views against the ground truth: the reference direction, its two sides, and the back. The side and back views thus test the completed surface, not only the reference view. CLIP-I compares the reference-view rendering with the input image. FD$_{\mathrm{DINOv3}}$, FD$_{\mathrm{CLIP}}$, FID, and CMMD compare the set of 16 rendered views of generated assets with the corresponding ground-truth set within each dataset. Reconstruction and CLIP-I measure fidelity to a specific object; the distribution metrics measure the visual quality of the generated population.

\noindent\textbf{Baselines.}
We use the authors' released checkpoints and implementations for the open-source baselines, following their inference recipes with method-specific sampling, resolution, and texture-export settings. Integration includes input-format conversion and repairs to mesh preprocessing, UV construction, and export. For given-geometry texturing, each method receives the designated reference image and GT mesh with its original materials removed, using the method's texturing interface; the reference camera is supplied where the interface supports it. No GT material maps or additional GT images are provided. For image-to-3D generation, each method receives the same single reference image. Pixal3D$^\star$ and \method{} share predicted geometry, as specified in Sec.~\ref{sec:image-to-3d}.

Exported assets are converted from their method-specific coordinate frames for evaluation; similarity alignment is applied where indicated by $\dagger$ in the tables. Rendering and scoring follow the shared evaluation protocol. TEXTRIX~\citep{zeng2025textrix} was not included in the quantitative comparison because its official implementation and pretrained weights were unavailable when our experiments were conducted. The same availability constraint applied to NaTex, LaFiTe, and MatMart~\citep{lai2026natex,chen2026lafite,wu2026matmart}. We discuss these methods in Sec.~\ref{sec:related}, but do not mix their reported scores with our results because the datasets and evaluation protocols differ.

\subsection{Commercial Comparisons}
\label{app:commercial-protocol}
Commercial systems do not disclose their training data, so overlap with our ground-truth benchmarks cannot be ruled out. We therefore evaluate \method{} without additional fine-tuning on a separate set of 45 images, comprising real photographs sourced online or contributed by users and synthetic images generated by image generation models such as GPT-Image-2. We compare end-to-end with the five commercial systems in Table~\ref{tab:commercial}: Hunyuan3D (v3.1), Hyper3D (Rodin-2.5 Extremely High), Tripo (H3.1), Meshy (v7.1), and Hi3D (v3.0), using their most advanced configurations and accessed after September~24, 2026. Hyper3D is listed as Rodin in the table.

We use two-view conditioning only for commercial comparisons, combining the reference image with a back view synthesized from it through the interface in Appendix~\ref{app:texture-editing}. We submit only the original reference image to each commercial system. Auxiliary-view synthesis is also supported by commercial workflows such as Meshy's multi-view mode, which can generate missing views from a single image~\citep{meshy2026multiview}.

The 3D generator receives projected PBR observations rather than photographs, so generalization to real inputs depends on the fine-tuned 2D material estimator and accurate placement of the surface observations. In the absence of ground-truth geometry and cameras, paired reconstruction metrics are unavailable. We instead assess three criteria: alignment with the input image, including text, logos, and color layout; PBR plausibility; and overall visual quality.

We collected 512 valid pairwise evaluations, each comprising one participant's judgments on all three criteria for \method{} versus one commercial system on a given input case. Each participant evaluated at most one model pair per case. For each of the five commercial systems and each criterion, Table~\ref{tab:commercial} reports \method{}'s wins divided by the total number of non-tied votes, pooled across evaluations.

Qwen3.8-27B assigns each result a score from 0 to 10 using the rendered outputs. Table~\ref{tab:commercial} reports both human preferences and VLM ratings. Figure~\ref{fig:commercial} shows how \method{} preserves reference details and coherently completes unseen surfaces using only public training data.

\section{2D Material Evaluation}
\label{app:quantitative}

\subsection{2D Material Prediction}
\label{app:2d-material}

We evaluate $F_{\mathrm{2D}}$ directly in image space, before 3D projection, sparse encoding, or completion. The evaluation includes all 100 objects of DTC-Tex and all 110 objects of Complex-Tex, using each object's designated reference view. Predictions use one deterministic step at native $1024\times1024$ resolution, CFG~1, and seed~42. All 210 cases complete successfully, with no exclusions. The auxiliary normal output is generated but is not scored here.

\noindent\textbf{Targets and metric domains.}
We reuse the benchmark's existing reference-view PNG targets without resizing or inverse display/gamma conversion. Each image is independently composited onto black using its own alpha, when present. Albedo PSNR and albedo, metallic, and roughness MAE use the GT foreground ($\alpha>0.5$). Albedo full-image PSNR uses the mean squared error over all RGB pixels with unit data range. Albedo SSIM uses the full image with a $7\times7$ window and unit data range; VGG-LPIPS also uses the full image. None of the full-image metrics masks prediction pixels with the GT foreground. Metallic and roughness predictions are read from the red and green channels of the MRA output. Metrics are computed per object and averaged separately within each dataset.

Metrics are computed in the stored PNG encoding; metallic and roughness MAE therefore measure encoded-value differences rather than physical-linear material error.

\begin{table}[!t]
\caption{Standalone 2D material prediction. Metrics are averaged over objects in the benchmark PNG code domain; metallic and roughness MAE do not measure physical-linear error.}
\label{tab:supp-2d-material}
\centering
\small
\resizebox{\linewidth}{!}{%
\begin{tabular}{lrrrrrrrrr}
\toprule
Dataset & $n$ & Alb. PSNR$_{\mathrm{fg}}\uparrow$ & Alb. PSNR$_{\mathrm{full}}\uparrow$ & SSIM$_{\mathrm{full}}\uparrow$ & LPIPS$_{\mathrm{full}}\downarrow$ & Alb. MAE$_{\mathrm{fg}}\downarrow$ & M MAE$_{\mathrm{fg}}\downarrow$ & R MAE$_{\mathrm{fg}}\downarrow$ & $F_{\mathrm{2D}}$ (s)$\downarrow$ \\
\midrule
\splitname{DTC-Tex} & 100 & 20.19 & 28.25 & 0.9678 & 0.0331 & 0.0872 & 0.0267 & 0.2540 & 1.682 \\
\splitname{Complex-Tex} & 110 & 19.44 & 28.19 & 0.9596 & 0.0427 & 0.1011 & 0.1167 & 0.2025 & 1.684 \\
\bottomrule
\end{tabular}}
\end{table}

\noindent\textbf{Results.}
The measured foreground PSNR is approximately 19--20\,dB, so these results do not establish near-ground-truth observation accuracy. Full-image PSNR is 28.25 and 28.19\,dB on DTC-Tex and Complex-Tex, respectively. Like full-image SSIM and LPIPS, it includes the shared background and should be interpreted separately from foreground PSNR. This experiment measures standalone $F_{\mathrm{2D}}$; the effect of its errors on 3D generation is compared in Sec.~\ref{sec:ablations}, with settings provided in Appendix~\ref{app:ablations}.

\subsection{\texorpdfstring{$F_{\mathrm{2D}}$}{F2D} Runtime}
\label{app:material-estimator-runtime}

Each request runs on one NVIDIA H800. After two warmup calls per worker, synchronized wall time covers input loading, conditioning, the single DiT call, and VAE/PIL decoding; model initialization and PNG writing are excluded. The mean isolated DiT latency is 1.205\,s on both datasets, and peak allocated memory is 66.25\,GiB. Four independent workers produce the cohort, with one GPU per request.

\noindent\textbf{Repeated DiT forward timing.}
We additionally measure the single DiT forward pass with the model loaded once and kept resident on one NVIDIA H800. After 10 excluded warmup calls, we time 30 repetitions of one reference image and 30 calls cycling through six reference images (three each from DTC-Tex and Complex-Tex). CUDA-event mean forward times are 1.1990\,s and 1.1987\,s, respectively, rounded to 1.20\,s in the main text. Each call jointly predicts albedo, metallic--roughness, and normal outputs at $1024\times1024$ with BF16, one inference step, CFG~1, and seed~42. Timing covers only the DiT forward pass, excluding model loading, conditioning, VAE encoding/decoding, and file I/O.

\section{Ablation and Observation Analyses}
\label{app:ablations}

\subsection{Ablation Configurations}

Table~\ref{tab:ablations} reports the architecture and material-content ablations. Table~\ref{tab:supp-ablation-settings} specifies the observation source and update rule for each row. The architecture variants use the same training data, initialization protocol, optimization settings, training budget, and checkpoint-selection rule. All variants use the given-geometry protocol of Sec.~\ref{sec:pbr-generation}: GT geometry and cameras, matched valid observation support and sampling seed, and 25 Euler steps without classifier-free guidance. Scores average 16 views per object on each dataset; shaded and albedo PSNR$_{\mathrm{fg}}$ use the GT mask, and MR MAE averages metallic and roughness errors on the GT--prediction foreground intersection.

\noindent\textbf{How observations are used.}
The architecture rows compare the conditioning designs as a whole.

\noindent\textbf{What observations carry.}
The material-content comparison retains predicted albedo and replaces only metallic and roughness with fixed or dataset-mean constants. Ours uses all predicted PBR channels. Oracle PBR replaces the local observations with GT materials and is reported separately as a material-quality reference. The standalone image-space accuracy of $F_{\mathrm{2D}}$ is reported in Appendix~\ref{app:2d-material}.

\subsection{Partial and Complete Encoding: Diagnostic Protocol}
\label{app:observation-analysis}
The results and reconstruction examples appear in Sec.~\ref{sec:observation-analysis}.

\noindent\textbf{Diagnostic protocol.}
Figure~\ref{fig:observation-analysis} reports the paired GT partial-versus-complete encoding analysis and three illustrative material reconstructions. The diagnostic uses each object's designated reference view, a two-pixel foreground erosion, and a $3\times3$ depth-discontinuity filter with relative threshold 0.04; projection IoU must be at least 0.90. The latent summary weights objects equally and uses 10,000 object-bootstrap resamples. The decoded examples are the objects nearest the 10th, 50th, and 90th percentiles of object-level latent RMS. Their $128\times128$ crops maximize GT albedo gradient on a 16-pixel candidate grid with at least 80\% valid pixels. Material values are converted to $[0,1]$ and clamped as in the production decoder path; PSNR uses the same retained GT surface voxels for both reconstructions. Geometry support is supplied by each encoder's spatial correspondence cache, so coordinate coverage is not evidence of independent geometry recovery.

\subsection{Observed-Support Encoding versus Zero Filling}
\label{app:zero-filling}
\begin{figure}[!t]
  \centering
  \includegraphics[width=\linewidth]{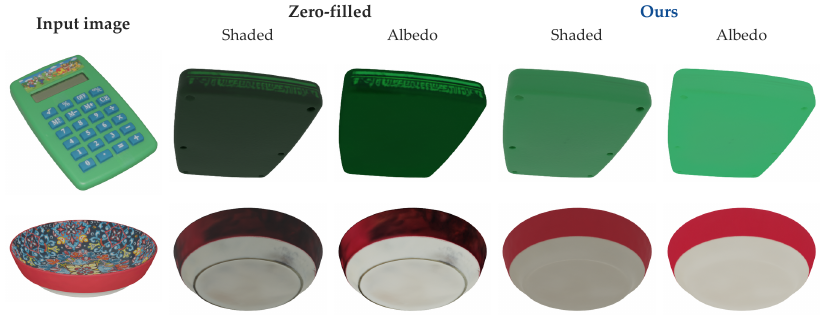}
  \caption{\textbf{Dark bias on unseen surfaces with zero-filled observations.} The calculator and dish are shown with their input images, followed by shaded and albedo renderings from the zero-filled model and \method{}. The zero-filled results darken the unseen surfaces; \method{} retains the corresponding green and red-and-white materials. Each comparison uses the same geometry and perspective view.}
  \label{fig:zero-filled-partial}
\end{figure}

As discussed in Sec.~\ref{sec:alignment}, zero-padding unobserved regions assigns the missing values concrete material semantics, such as black albedo. In the zero-filled formulation, we observed a pronounced dark bias on unseen surfaces. Figure~\ref{fig:zero-filled-partial} illustrates this behavior: the back of the calculator becomes nearly black and retains spurious front-side characters, while the dish develops dark mottling on its red side and white underside. With observed-support-only encoding, \method{} instead produces a green calculator back and preserves the dish's red-and-white color regions.

\subsection{Early Learning with Native PBR Latents}
\label{app:native-pbr-learning}

The native PBR latent is an \emph{in-domain} observation representation: visible material evidence is encoded by the same pretrained material VAE that defines the generator's output space. We examine how quickly the generator learns to recover reference-view appearance from this representation, using the official Pixal3D$^\star$~\citep{li2026pixal3d} implementation as an image-feature-conditioned comparison.

Both models are trained on the same training set and initialized from the same pretrained TRELLIS.2 denoiser with their respective architectural adaptations and fresh optimizers. TeaPot is used only for evaluation, not for single-object fitting. We evaluate all 16 raw checkpoints from step~0 to~1500 at intervals of 100 on its designated front view; there are no failed or excluded checkpoints. Our model receives ground-truth albedo, metallic, roughness, and opacity through native partial-material encoding, while Pixal3D$^\star$ receives the corresponding front-lit RGB image through its official image-feature projection interface. This oracle-material setting examines the 3D model's ability to learn from native observations independently of $F_{\mathrm{2D}}$ prediction errors. The two conditioning designs retain their respective architectures and optimization recipes.

\begin{table}[!t]
  \caption{Configurations for the six rows of Table~\ref{tab:ablations}. All observation conditions are encoded with the pretrained material VAE. $\widehat{\mathbf a}$, $\widehat m$, and $\widehat r$ denote predicted albedo, metallic, and roughness.}
  \label{tab:supp-ablation-settings}
  \centering
  \small
  \setlength{\tabcolsep}{4pt}
  \begin{tabular}{@{}p{0.22\linewidth}p{0.30\linewidth}p{0.42\linewidth}@{}}
    \toprule
    Variant & Material observations & Conditioning architecture \\
    \midrule
    Residual injection & Predicted PBR & Observation latents added as layerwise residuals. \\
    Joint-update attention & Predicted PBR & Observation states updated in selected layers; all generation states updated in every layer. \\
    Fixed MR & $\widehat{\mathbf a}$; $m=0$, $r=0.5$ & Read-only observations; interleaved generation updates. \\
    Dataset-mean MR & $\widehat{\mathbf a}$; dataset-mean $m,r$ & Same architecture as Ours. \\
    \midrule
    \rowcolor{supraTableWash}
    Ours & Predicted PBR & Read-only observations; alternating unobserved-cell and full-surface updates. \\
    Oracle PBR (GT) & GT PBR & Reported separately as a material-quality reference. \\
    \bottomrule
  \end{tabular}
\end{table}

\begin{figure}[!t]
  \centering
  \includegraphics[width=0.90\linewidth]{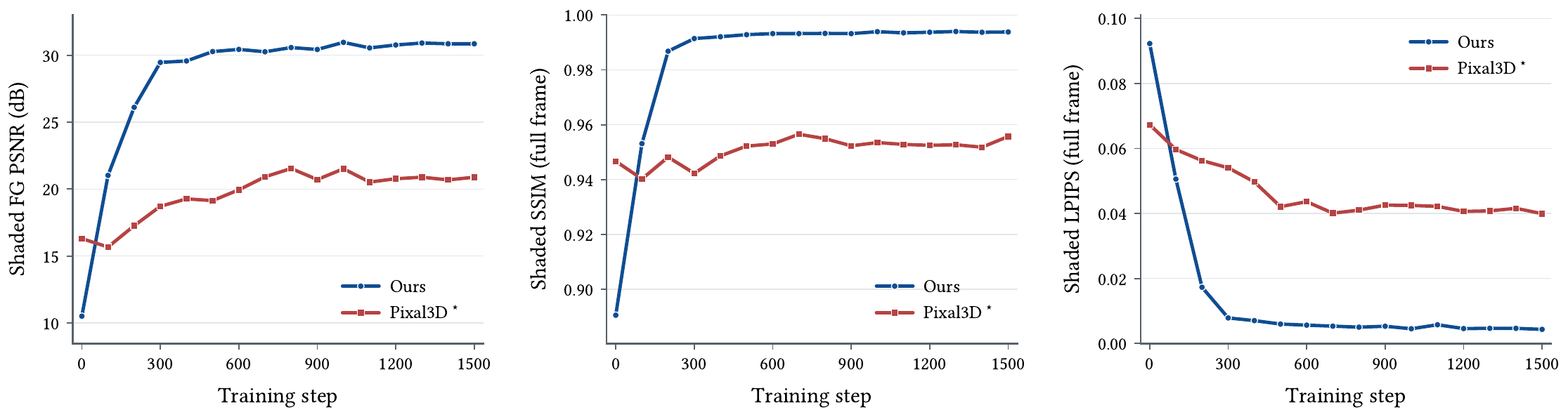}
  \caption{\textbf{Early reference-view learning on TeaPot during full-data training.} Ours uses GT PBR observations; Pixal3D$^\star$ uses the corresponding shaded RGB image. Each point is an evaluated checkpoint. Most of our model's improvement occurs within the first 300 steps.}
  \label{fig:native-pbr-learning}
\end{figure}

All checkpoints use the same GT geometry and camera, 25 Euler sampling steps, seed~42, and time rescaling factor~3, with no reprojection repainting or material postprocessing. We render with Blender~3.0.1 using Filmic color management and 32 samples at $1024\times1024$, then resize to $512\times512$ with Lanczos for scoring. Foreground PSNR uses the same 31,016 GT-mask pixels without erosion or intersection with the prediction mask. SSIM and VGG-LPIPS use the full image, including the background.

\begin{figure}[!t]
  \centering
  \includegraphics[width=0.90\linewidth]{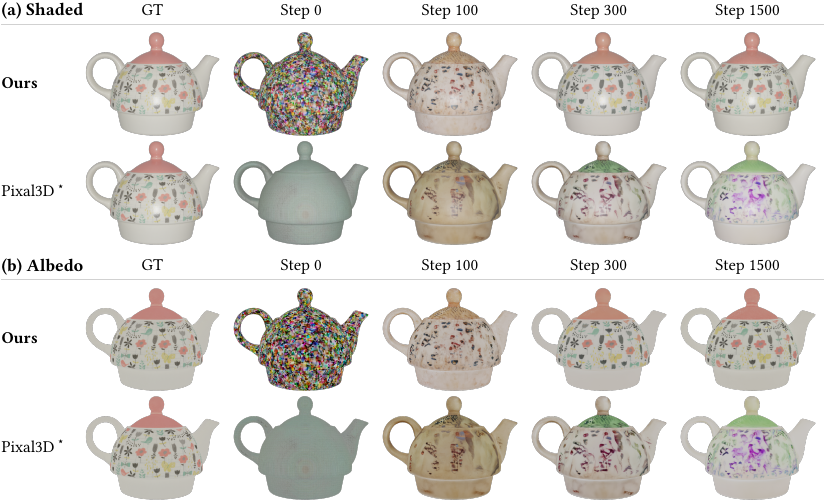}
  \caption{\textbf{Reference-view appearance during early training.} Renderings correspond to the experiment in Fig.~\ref{fig:native-pbr-learning}; our model recovers the reference colors and floral patterns within 300 steps.}
  \label{fig:native-pbr-learning-qualitative}
\end{figure}

With GT PBR observations, our model reaches 29.46\,dB shaded foreground PSNR on the TeaPot reference view after 300 steps, compared with 18.72\,dB for Pixal3D$^\star$ under shaded RGB conditioning. At step~1500, the scores are 30.86 versus 20.89\,dB, SSIM is 0.9938 versus 0.9557, and LPIPS is 0.00436 versus 0.03996. Figure~\ref{fig:native-pbr-learning-qualitative} shows the early recovery of reference colors and floral patterns. This single-view diagnostic supports the learnability of native observations under oracle-material conditioning; it does not isolate conditioning from the models' architectural and optimization differences. Full-surface and dataset-level quality are evaluated separately in the main experiments.

\section{Additional Qualitative Comparisons}
\label{app:extensions}

\subsection{Given-Geometry Texturing}
\begin{figure}[!t]
  \centering
  \includegraphics[width=\linewidth,height=\textheight,keepaspectratio,page=1]{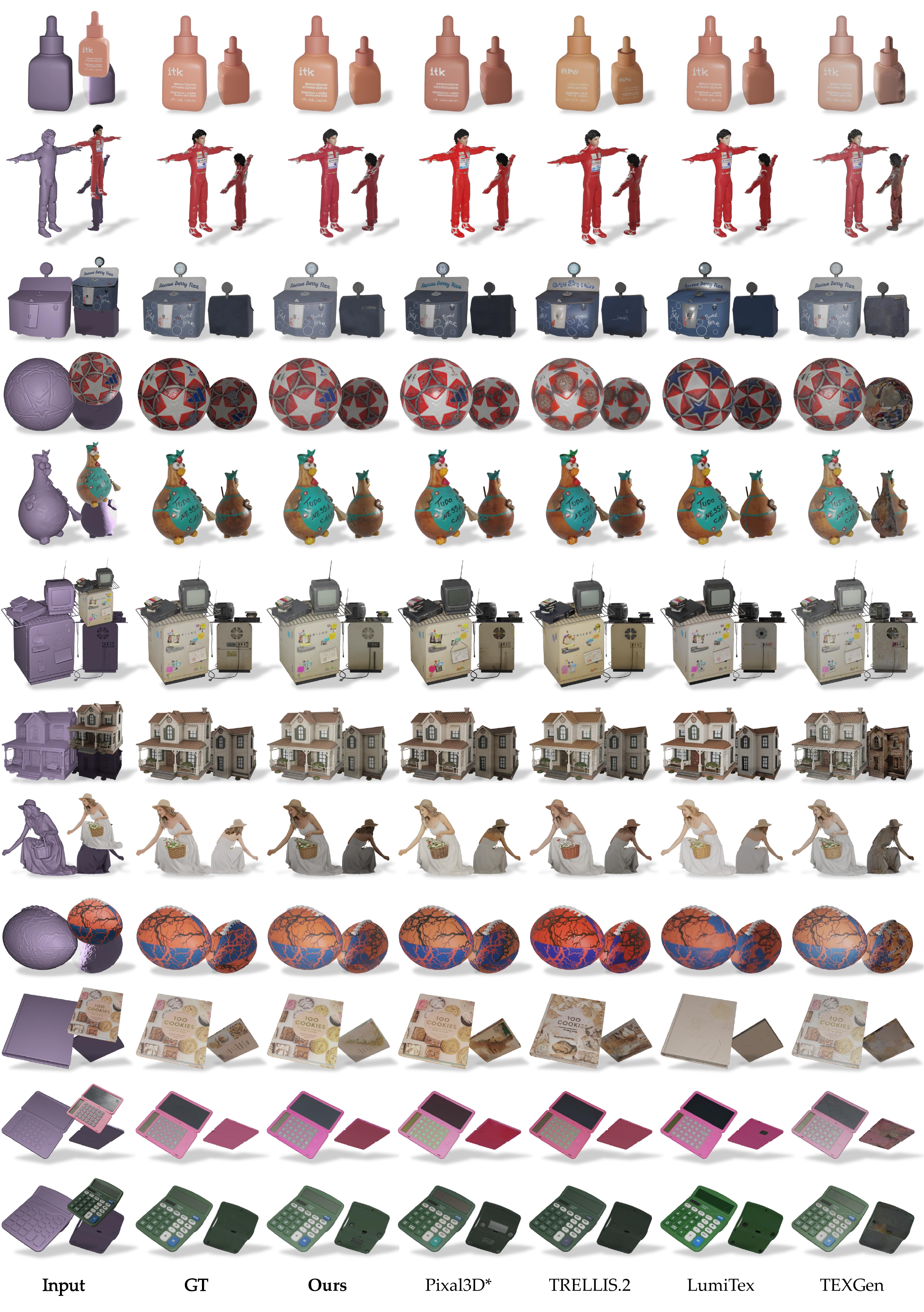}
  \caption{Twelve given-geometry comparisons. Novel views show the completed appearance.}
  \label{fig:supp-tracka-qualitative}
\end{figure}

Figure~\ref{fig:supp-tracka-qualitative} provides additional given-geometry comparisons across twelve cases.

\subsection{End-to-End Generation}
\begin{figure}[!t]
  \centering
  \includegraphics[width=\linewidth,page=1]{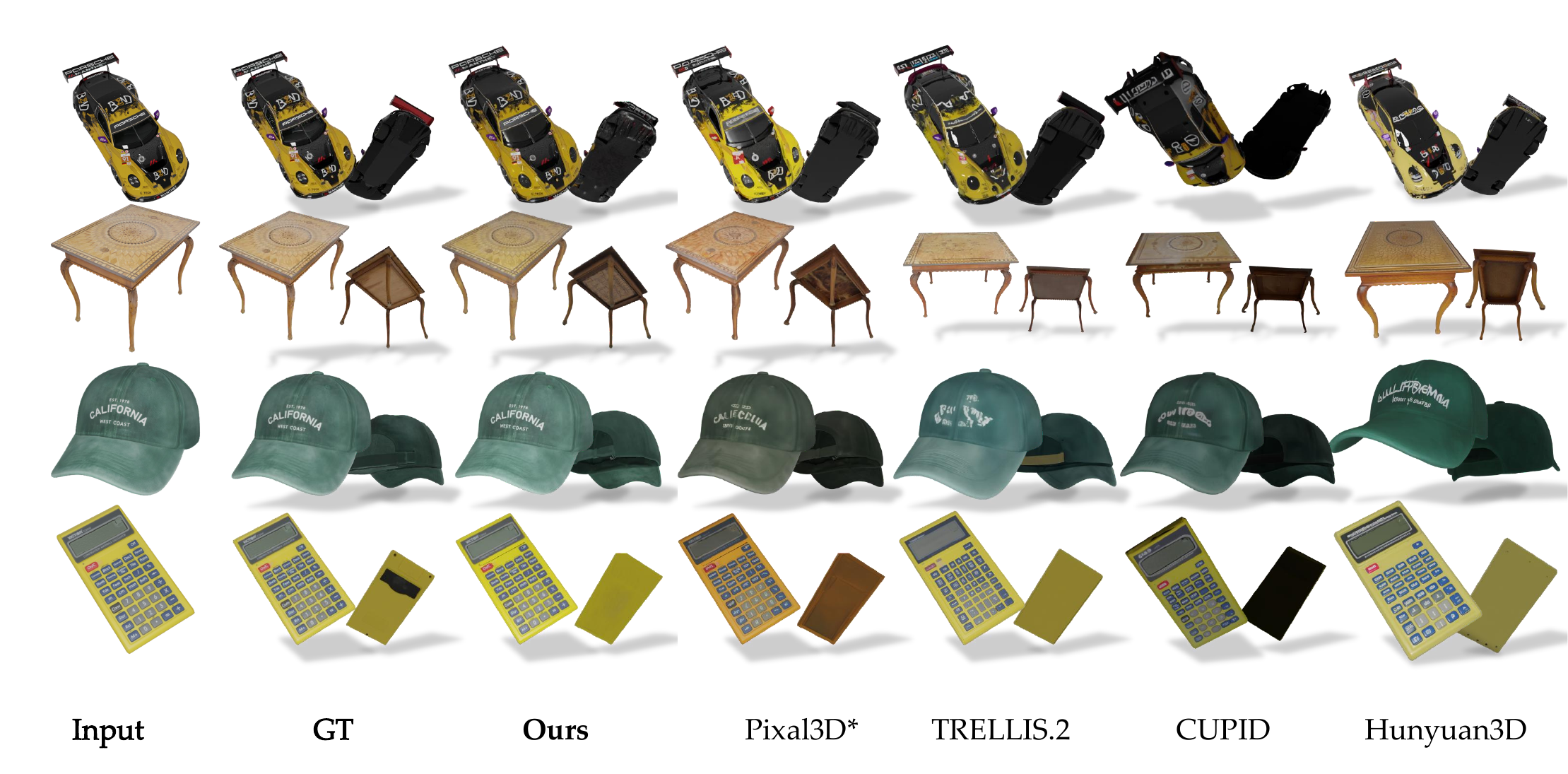}
  \caption{Qualitative comparison of end-to-end single-image asset generation (Table~\ref{tab:image-to-3d}) across four cases. All methods use the same reference image. Pixal3D$^\star$ and \method{} share predicted geometry; other baselines use their own geometry pipelines. For each object we show the input, the reference-view rendering, and a novel view. Reference views assess text and repeated patterns; novel views assess the continuity of the completed appearance.}
  \label{fig:image-to-3d-qualitative}
\end{figure}

With predicted geometry, the question in the images is whether local details still land on the right surface. Figure~\ref{fig:image-to-3d-qualitative} places each input image alongside reference-view and novel-view renderings from each method to compare how these details carry over to the generated asset. Reference views highlight text, logos, repeated patterns, and color boundaries; novel views show the continuity of the completed appearance.

\subsection{Commercial Comparisons}
\begin{figure}[!t]
  \centering
  \includegraphics[width=\linewidth]{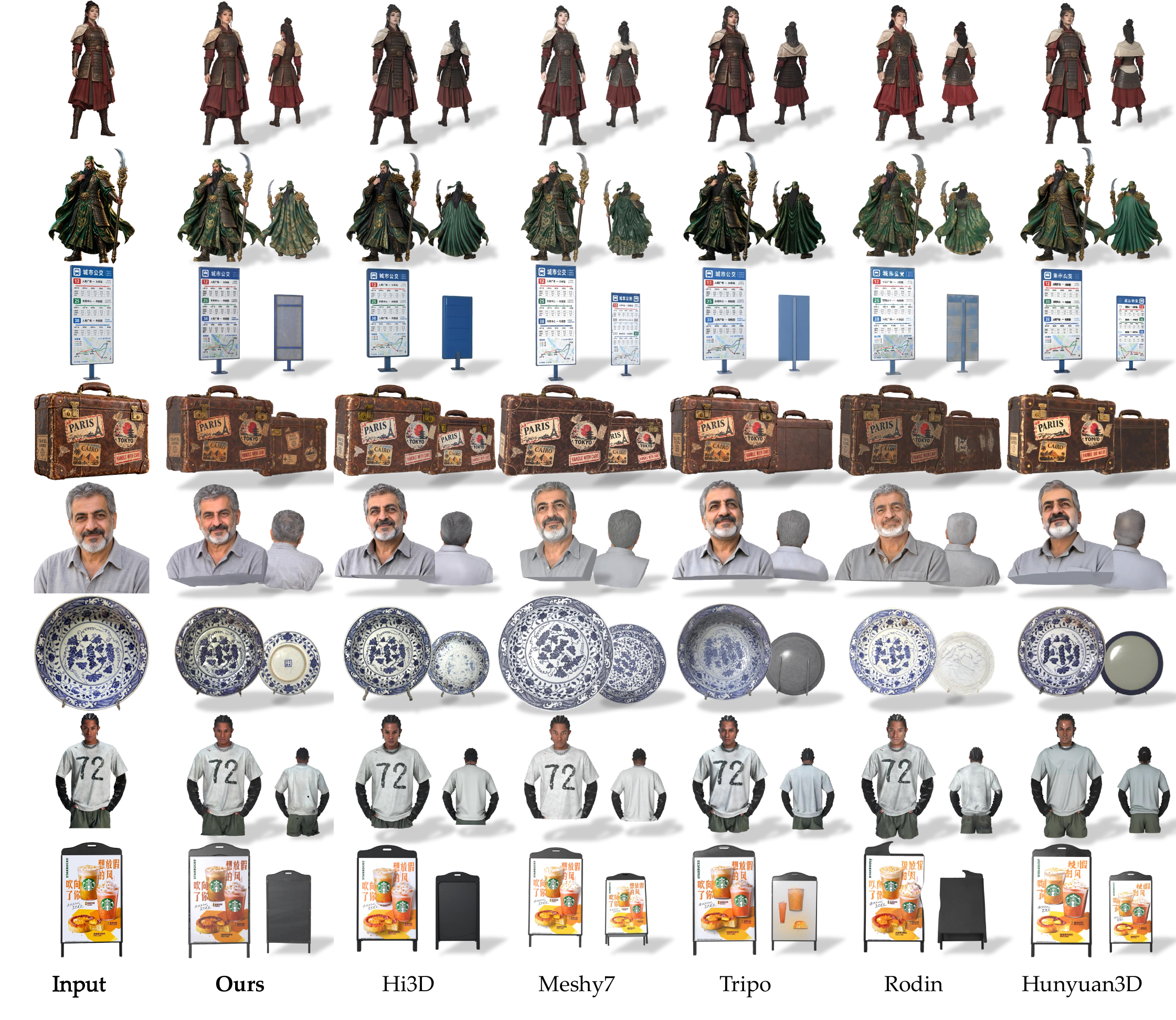}
  \caption{Additional comparisons with commercial models across eight cases. Each row shows the input image and front/back renderings from Hunyuan3D, Rodin, Tripo, Meshy, Hi3D, and \method{}. Model versions and evaluation settings are given in Appendix~\ref{app:commercial-protocol}.}
  \label{fig:supp-commercial}
\end{figure}

Figure~\ref{fig:supp-commercial} complements Figure~\ref{fig:commercial} with eight additional cases under the protocol in Appendix~\ref{app:commercial-protocol}. Front views show the reproduction of text, patterns, and color layout from the input, while back views show the completion of unseen surfaces.

\begin{figure}[!t]
  \centering
  \includegraphics[width=\linewidth]{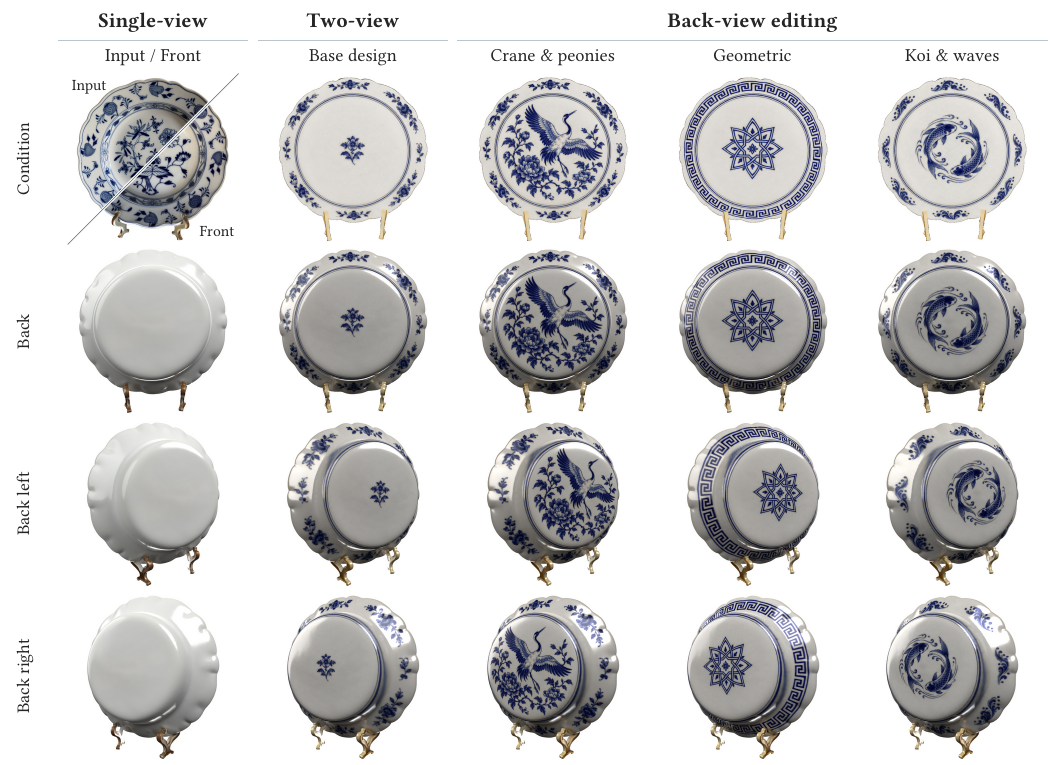}
  \caption{Single-view generation, two-view fusion, and back-view editing. The first condition panel shows the input photograph (upper left) and generated front (lower right); other panels show designed back-view materials.}
  \label{fig:texture-editing-plate}
\end{figure}

\section{Additional Views and Texture Editing}
\label{app:editing-limitations}

\subsection{Single-View Generation and Back-View Editing}
\label{app:texture-editing}
\label{sec:extensions}

Because any image attached to the surface becomes an anchor (Sec.~\ref{sec:alignment}), the same generator accepts a second view or an edited view without retraining: a 2D editing model can supply appearance for surfaces the reference never shows, or change the appearance of surfaces it does. Both put the unseen surface under user control: the single-view results of Secs.~\ref{sec:pbr-generation} and~\ref{sec:image-to-3d} show what the generator completes on its own, and the extensions show what it produces when instructed. We demonstrate both with the same 3D generator and no additional 3D training.

\noindent\textbf{Synthetic back-view conditions.}
From the same reference photograph, we synthesize a geometry-consistent back view and encode it as an additional material condition. Specifically, we select a viewpoint with minimal visual overlap relative to the reference, typically the opposite back side, render its surface normal map, and synthesize its shaded appearance via an MLLM, such as GPT-image, guided by the reference image, the rendered normal map, and optional user prompts. Both shaded views are translated into PBR maps, aggregated in 3D coordinate space via projection, and encoded into unified anchors $O$ for full-surface completion. We use this two-view variant for comparisons with commercial systems; the benchmark comparisons in Tables~\ref{tab:pbr-generation} and~\ref{tab:image-to-3d} use the single-view model.

\noindent\textbf{Texture editing.}
Editing the observation condition transfers image edits to the 3D surface. We use the image editing model to modify the front or back view, rebuild that view's observations, and regenerate the complete material. The visualizations show how the requested edit appears on the corresponding surface and how the edited region stays consistent with the rest of the asset.

\noindent\textbf{Porcelain-plate example.}
Figure~\ref{fig:texture-editing-plate} demonstrates single-view generation, two-view fusion, and image-space material editing on a porcelain plate. Starting from the same front photograph and source geometry, the single-view model completes the unobserved back. A geometry-guided 2D design supplies a second material observation; editing that design produces crane-and-peony, geometric, and koi variants without additional 3D training. The synthetic back is a design hypothesis rather than a second photograph of the real object.

We keep the source geometry, front material observation, model checkpoint, and sampling settings fixed. Front and back material observations are combined on the shared surface before native material encoding. The diagonal input--render comparison shows reference-view fidelity, while the back and opposite rear oblique views show how the edited motifs are carried onto the complete asset.

\FloatBarrier
\section{Notation Reference}
\label{app:notation}

Table~\ref{tab:supp-notation} summarizes the notation used in the main paper and supplement.

\begingroup
\small
\begin{longtable}{p{0.14\textwidth}p{0.24\textwidth}p{0.52\textwidth}}
\caption{Notation and symbol reference.}\label{tab:supp-notation}\\
\toprule
\textbf{Symbol} & \textbf{Domain / Dimensions} & \textbf{Physical or Mathematical Interpretation} \\
\midrule
\endfirsthead
\multicolumn{3}{l}{\small\textit{Table~\thetable{} (continued).}}\\
\toprule
\textbf{Symbol} & \textbf{Domain / Dimensions} & \textbf{Physical or Mathematical Interpretation} \\
\midrule
\endhead
\midrule
\multicolumn{3}{r}{\small\textit{Continued on the next page.}}\\
\endfoot
\bottomrule
\endlastfoot
\multicolumn{3}{l}{\textit{Inputs and Geometry Foundation}} \\*
$I$ & $\mathbb{R}^{H \times W \times 3}$ & Input reference photograph in RGB space \\
$\mathcal M$ & 3D Surface & Given or upstream-predicted object surface mesh \\
$\pi$ & Projection operator & Camera intrinsic and extrinsic projection mapping 3D to 2D \\
$\mathbf p \in \mathcal M$ & $\mathbb{R}^3$ & Continuous 3D coordinate on the mesh surface \\
$\mathbf x_s$ & $\mathbb{R}^3$ & Continuous 3D centroid of sparse latent cell $s$ \\
\midrule
\multicolumn{3}{l}{\textit{PBR Materials and 2D Translation}} \\*
$T(\mathbf p)$ & $\mathbb{R}^3 \times [0, 1]^3$ & PBR material tuple $(\mathbf a(\mathbf p), m(\mathbf p), r(\mathbf p), \alpha(\mathbf p))$ at surface location $\mathbf p$ \\
$\mathbf a, m, r, \alpha$ & $\mathbb{R}^3, [0, 1], [0, 1], [0, 1]$ & Albedo, metallic, roughness, and opacity physical channels \\
$F_{\mathrm{2D}}$ & Modality-DiT & MLLM-based 2D image editing model for reference-view PBR estimation \\
$\widehat T_{\mathrm{2D}}$ & $\mathbb{R}^{1024 \times 1024 \times 5}$ & Predicted physical maps $(\widehat{\mathbf a}, \widehat m, \widehat r)$; MR is packed with a constant channel for the image VAE \\
$\mathbf u$ & $\mathbb{Z}^2$ & 2D pixel coordinate in reference image and PBR maps \\
\midrule
\multicolumn{3}{l}{\textit{Sparse Coordinate Space and 3D VAE Latents}} \\*
$\mathcal S$ & Coordinate set ($|\mathcal S| = N$) & Set of active sparse voxel coordinates on the surface of $\mathcal M$ \\
$G = \{g_s\}_{s\in\mathcal S}$ & $\mathbb{R}^{N \times 32}$ & Full-surface geometry latents produced by shape VAE \\
$Z = \{z_s\}_{s\in\mathcal S}$ & $\mathbb{R}^{N \times 32}$ & Complete full-surface material latents (locked target representation) \\
$E_{\mathrm{mat}}, D_{\mathrm{mat}}$ & Neural networks & Pretrained 3D sparse material VAE encoder and decoder \\
$\mathcal P$ & Sparse attributes & Full-surface PBR attribute voxels (input to $E_{\mathrm{mat}}$ during ground-truth pretraining) \\
\midrule
\multicolumn{3}{l}{\textit{Native Observation Encoding and Spatial Metadata}} \\*
$\mathcal H_{\mathrm{valid}}$ & $\{(\mathbf u, \mathbf p)\}$ & Valid pixel-surface correspondence pairs retained for observation encoding \\
$\mathcal P_o$ & Sparse attributes & Voxelized partial PBR attributes supported strictly on visible surface \\
$\mathcal S_o \subseteq \mathcal S$ & Coordinate subset ($|\mathcal S_o| = M$) & Coordinate indices of latent cells that received valid observations \\
$z_s^o$ & $\mathbb{R}^{32}$ & Native observation latent at cell $s\in\mathcal S_o$ encoded by $E_{\mathrm{mat}}$ \\
$\mathcal U = \mathcal S \setminus \mathcal S_o$ & Coordinate subset & Indices of unobserved latent cells required to be completed \\
$\mathcal V_s, \mathcal V_o$ & Discrete voxel sets & Surface voxels in latent cell $s$, and the deduplicated observed surface voxels \\
$\rho_s \in (0, 1]$ & Scalar ratio & Observation coverage fraction $|\mathcal V_s \cap \mathcal V_o| / |\mathcal V_s|$ of cell $s$ \\
$d_s \in \mathbb{Z}_{\ge -1}$ & Graph distance & 6-neighbor lattice geodesic distance from cell $s$ to $\mathcal U$ ($-1$ if unreachable) \\
$\mathcal C_o$ & Structured set & Native observation representation $\{(s, z_s^o, \rho_s, d_s) \mid s\in\mathcal S_o\}$ \\
\midrule
\multicolumn{3}{l}{\textit{Observation-Guided Completion and Interleaved Flow Transformer}} \\*
$t \in [0, 1]$ & Continuous scalar & Flow matching time ($t=1$: Gaussian noise; $t=0$: material endpoint with residual noise $\sigma_{\min}\epsilon$) \\
$Z_t$ & $\mathbb{R}^{N \times 32}$ & Noisy full-surface material latents along probability trajectory at time $t$ \\
$X = \{X_s^{(\ell)}\}$ & $\mathbb{R}^{N \times 1536}$ & Dynamic full-surface generation stream hidden states across blocks $\ell$ \\
$O = \{O_s\}$ & $\mathbb{R}^{M \times 1536}$ & Static read-only observation stream memory (fixed anchor representation) \\
$\ell \in \{0, \ldots, L-1\}$ & Integer ($L=60$) & Transformer block index \\
$\mathcal J_\ell$ & Coordinate subset & Query coordinates for block $\ell$ ($\mathcal J_\ell = \mathcal U$ for even $\ell$, $\mathcal J_\ell = \mathcal S$ for odd $\ell$) \\
$W_{\mathrm{in}}$, $W_{\mathrm{meta}}$, $b_o$ & Learnable parameters & Shared material-geometry projection, metadata projection, and observation-role bias \\
$Q_\ell, K_\ell, V_\ell$ & Hidden tensors & Cross-stream attention queries (from $X_{\mathcal J_\ell}$), keys and values (from $[O; X]$) \\
$d_h=128$ & Per-head dimension & Attention head width ($1536/12$) \\
$v^*, v_\theta$ & $\mathbb{R}^{N \times 32}$ & Ground-truth flow target velocity and predicted full-surface velocity field \\
\end{longtable}
\endgroup